\documentclass[preprint,12pt]{elsarticle}

\usepackage[cmex10]{amsmath}
\usepackage{amsthm}

\usepackage{algorithmic}

\usepackage{array}

\usepackage[caption=false,font=normalsize,labelfont=sf,textfont=sf]{subfig}
\usepackage{url}
\journal{Signal Processing: Image Communication}

\usepackage[table,xcdraw]{xcolor}

\usepackage{color}

\newcommand{\etal}[1]{\textit{#1~et.al.}}

\usepackage{soul}

\begin{document}
%
\title{}

\newif\iffinal
\finaltrue
\newcommand{\jemsid}{99999}


\iffinal



%

\begin{frontmatter}
\title{Exposing Computer Generated Images by Using\\
Deep Convolutional Neural Networks}


\author[label1]{Edmar R. S. de Rezende}
\author[label1]{Guilherme C. S. Ruppert} 
\author[label1]{Ant\^{o}nio The\'{o}philo}
\author[label2]{Tiago Carvalho}
\address[label1]{CTI Renato Archer, Campinas-SP, Brazil 13069-901}
\address[label2]{Federal Institute of S\~{a}o Paulo (IFSP), Campinas-SP, Brazil 13069-901}

\begin{abstract}
The recent computer graphics developments have upraised the quality of the generated digital content, astonishing the most skeptical viewer. Games and movies have taken advantage of this fact but, at the same time, these advances have brought serious negative impacts like the ones yielded by fake images produced with malicious intents. Digital artists can compose artificial images capable of deceiving the great majority of people, turning this into a very dangerous weapon in a timespan currently know as ``Fake News/Post-Truth'' Era. In this work, we propose a new approach for dealing with the problem of detecting computer generated images, through the application of deep convolutional networks and transfer learning techniques. We start from Residual Networks and develop different models adapted to the binary problem of identifying if an image was or not computer generated. Differently from the current state-of-the-art approaches, we don't rely on hand-crafted features, but provide to the model the raw pixel information, achieving the same 0.97 of state-of-the-art methods with two main advantages: our methods show more stable results (depicted by lower variance) and eliminate the laborious and manual step of specialized features extraction and selection.

\end{abstract}


\end{frontmatter}

\section{Introduction}
\label{sec:intro}

The 2016 \textit{Global Games Market Report}\footnote{\url{https://goo.gl/xkWPon}} presented the economic potential of digital games, which traded more than 99.6 billions of dollars, an increment of 8\% when compared to the previous year. The growth in this market always pushes forward the quality and development of associated industries and technologies as, for example, computer graphics methods. These methods are essential to make games more realistic through high quality graphics.

Another entertainment field that takes advantage of advanced computer graphics methods is the movies industry. Thinking about realism, in the last years we've experienced huge steps towards a complete deceiving of our visual senses. Productions as \textit{Rogue One: A Star Wars Story} showed the potential of Computer Graphics (CG) characters construction, introducing in a live action movie characters entirely based on real actors.

The search for a perfect generation of digital scenarios, objects and even people is endless and recently reached an astonishing point, mostly helped with the latest advances of computing processing, in special the modern GPU cards (Graphics Processing Units). One current example of such an achievement was the digital reproduction of the actress Carrie Fisher in the last Star Wars movie~\footnote{\url{http://www.imdb.com/title/tt3748528/}}, with the same appearance of the beginning of her career in the 70's.

In spite of the safe and benign results of these advances, once the goal of perfect CG image generation is accomplished, some threats come along and introduce new challenges to other science areas as pointed out by~\citeauthor{holmes_etal}~\cite{holmes_etal}. One example of such a challenge is the identification if an image was a photo generated (PG - the one generated by a digital camera) or generated by CG methods. Figure~\ref{fig:intro}, shows an example of how difficult is to discern between PG and CG images.

\begin{figure}[!h]
  \centering
  \subfloat[PG]{\includegraphics[width=0.40\columnwidth]{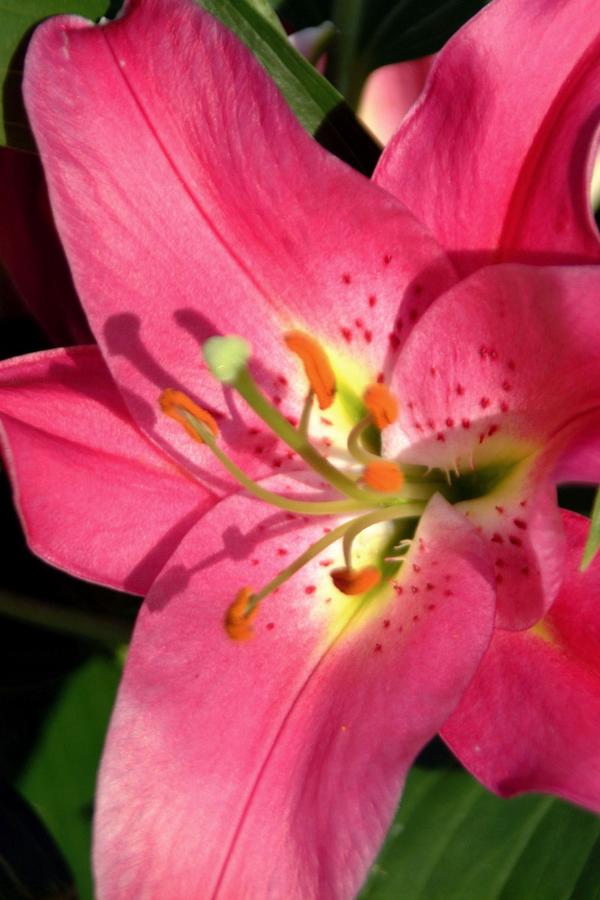}} \hspace*{0.1cm}
  \subfloat[CG]{\includegraphics[width=0.405\columnwidth]{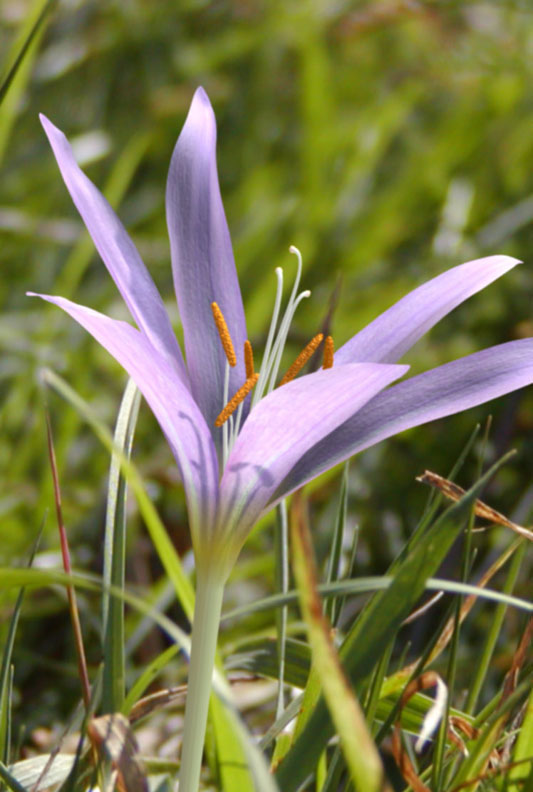}}
  \caption{Example of how challenging is to recognize PG and CG images by simple visual analysis.}
 \label{fig:intro}
\end{figure}

Recent studies showed how easy is to deceive people using images~\cite{SCHETINGER2017142}. In special, several examples of undesired situations can be described involving the CG images. Imagine, for example, a CG image depicting a terrorist execution of a kidnapped report spreading across the globe. Or another CG image of a rising politician, posted in social networks putting him in an embarrassing or criminal situation days before an election. We are living in what some are calling the Fake News/Post-Truth Era~\cite{schulten:fake:2017,keyes:post:2004}, where mass communication platforms (as social media networks) can be used to influence and deceive people~\cite{davey:fake:2017,shane:fake:2017}. If a well-crafted invented text can have a great impact on people's public opinion, imagine the effects of a CG image produced by a very (probably well-paid) skilled professional posted on social networks.

The distinction between PG and CG has even more complex legal implications when related to child pornography. In Brazil, any person who produces, reproduces, directs, takes pictures or records, in any way, scenes involving explicit sexual or pornographic act involving children or teenager, can be sued according to Brazilian Law 11,829 published on November 25\textsuperscript{th}, 2008. This legal process can result in 4 to 8 years in jail. This situation raises a fundamental legal and ethical question created by technology: What happens if the material is proved to be CG generated? Are the legal consequences the same?

The task of CG image and video detection was already studied and several Digital Forensics methods were proposed~\cite{Tokuda20131276,_dang_nguyen_etal_a_,_dang_nguyen_etal_b_,_farid_and_bravo_2012_}. However, the results are far from considering the problem as completely solved. Very often these methods are based on the discovery of inconsistencies in very specific situations, hindering their wide application. For example, \citeauthor{conotter14}~\cite{conotter14} developed a method based on blood flow information of CG constructed people in videos. In contrast, \citeauthor{Tokuda20131276}~\cite{Tokuda20131276} proposed a more generic method that applies machine learning techniques to solve the CG image identification task and is more similar with the one presented by this work.

The rise of Deep Neural Networks (DNN) in the past few years, presented a shift in classification process, specially in the feature engineering step of this process. Algorithms based on DNN have outperformed other approaches in image classification, becoming the standard approach for these tasks. They consist of learning algorithms with multiple levels, acting over the raw input (image pixels for example) transforming the representation at one level into a representation at a higher, slightly-more-abstract level~\cite{bengio2009learning,Goodfellow-et-al-2016,LeCun2015}. As this stack of layers gets bigger, more complex functions can be learned from data. Besides this power of representing more complicated mappings, a great advantage of DNN is that there is no need for human engineered features, with a general purpose algorithm learning direct from raw data.

In spite of the basic concepts of DNN being around for some decades, only now, with the plenty availability of data and the recent developments of GPU cards, DNN showed its full potential, specially in image classification challenges such as the ILSVRC (ImageNet Large Scale Visual Recognition Challenge)~\cite{russakovsky2015imagenet}. This highlighted the transition from hand-crafted features combined with shallow classifiers to deep classifiers acting directly on raw data as is the case of DNNs.

Since then, there is been a trend of, as deeper the model is, the better its performance and more difficult is the training process. This can be demonstrated by ImageNet challenge results. In 2012 the 8-layers AlexNet network~\cite{krizhevsky2012imagenet} astonished the machine learning community winning the challenge with a top-5 classification error rate of 16.4\% and a huge leap from the second place (this one using usually hand-crafted features and shallow classifiers). In 2014, two VGG DNN models (one with 16 layers and the other with 19 layers) got a top-5 classification error rate of 7.3\%~\cite{simonyan2014very} while GoogleNet with its 22 layers won the challenge with 6.7\% error rate~\cite{szegedy2015going}. Finally, in 2015 the Residual Network (ResNet) model, a DNN with 152 layers, achieved a top-5 classification error of 3.57\%~\cite{he2016deep}. Also, in 2015, for the first time, was presented a DNN technique capable of performing better than humans in image classification tests~\cite{he2015delving}.

This paper presents a novel approach for dealing with the task of detecting CG image generation. Two different models are developed, starting from the DNN ResNet with 50 layers (ResNet-50)~\cite{he2016deep} and adapting it to the binary problem of CG image detection. Applying concepts of transfer learning~\cite{yosinski2014transferable}, we were able to transfer the weights of ResNet-50 layers pre-trained on ImageNet dataset to our model, avoiding overfitting and achieving 97\% of accuracy without the burden of designing complex hand-craft features. To our best knowledge, this is the first work to propose applying DNN techniques to this problem, not requiring human experts to design features.

Regarding the actual state-of-the-art method for detection of CG image generation~\cite{Tokuda20131276}, the main contributions of this paper are: (1) the proposal of a new approach based on DNN and transfer learning techniques that achieves the same accuracy of 0.97 as state-of-the-art methods without the need for human level feature extraction; (2) the use of an extended dataset (more difficult for the task); (3) a more stable method proved by the lower variance results\footnote{All the artifacts (code and dataset) produced by this work will be available in case of paper acceptance.}; (4) evaluation of different kinds of classifiers in association with a DNN in order to find the best combination (features + classifier); (5) and a qualitative analysis of bottleneck features produced by ResNet-50 in CG image detection problem.

The text is structured in the following way: Section \ref{sec:relatedwork} briefly presents the main works in the Digital Forensics literature that deal with the problem of detecting CG image generation. Section~\ref{sec:proposedmethod} explains with details the proposed methodology while Section~\ref{sec:results} describes the main experiments conducted to validate the methodology and presents the achieved results, comparing with the state-of-the-art found on the literature. Lastly, Section~\ref{sec:conclusions} presents the main conclusions and some future research directions.

\section{Related Work}
\label{sec:relatedwork}

There are many works in the literature on the topic of distinguishing between CG and real images. \etal{Holmes}~\cite{holmes_etal} discusses the legal aspects related to the problem, specially for child pornography. The authors investigated the perception of humans exposed to this kind of image performing two experiments: (i) the first one in which a set of images (CG and real) are shown to untrained users, and (ii) the second where there was a previous training for users before showing the images. The experiment consisted in submitting each user to 60 pictures of people. The user were asked to identify the sex (man or woman) and if the image was real or generated by computer. In the first round of experiments, the untrained users achieved an accuracy around 50\% in CG image detection. In the second experiment, after a simple training, the users improved their accuracy at the task. Also, as the CG image quality improves, it becomes even easier to trick the perception of the user to distinguish real and CG images.

The work~\cite{farid-tr-04} discusses the US Supreme Court decision on not considering as a crime the computer generated child pornography. Also, the work presents techniques for image tampering and approaches to detect some kinds of image manipulation.

\etal{Conotter}~\cite{conotter14} proposed to use information associated with blood flow and perceptual details to detect computer generated people in videos. The method consists in evaluating small movements of cheeks and forehead to generate a distinguishable signal of CG and real images. This signal is more stable for real images, while CG images present many peaks.

Many methods based on machine learning have been proposed, which typically consists in extracting features and using a supervised learning classifier to identify patterns of CG or real images. \etal{Tokuda}~\cite{Tokuda20131276} proposes a method using fusion of many classifiers combined with a big number of feature extraction schemes, achieving a 97\% accuracy on his dataset (9700 images).

\etal{Tan}~\cite{Tan16} uses Local Ternary Patterns (LTP) for features extraction, and rely only on texture features to distinguish CG and real images. Experiments reveal that  the method achieves an accuracy of approximately 97\% in a dataset of 2200 images collected from different sources, as for example, the Columbia University natural image library~\cite{columbiadb},using a Support Vector Machine~(SVM)~\cite{Bishop:2006:PRM:1162264} classifier.

\section{Proposed Method}
\label{sec:proposedmethod}

The CG detection method proposed in this work relies upon a deep CNN architecture to classify each image from the dataset using the raw RGB pixels values as features, without the need for manual feature extraction. The deep CNN deployed is based on the ResNet-50 model~\cite{he2016deep} and the method uses transfer learning techniques~\cite{yosinski2014transferable}. All the pipeline of proposed method as fundamental concepts related with it will be explained in the next sections.

\subsection{Complete Model Architecture}

Our proposed deep CNN model uses transfer learning techniques, leveraging the outcomes of residual learning presented by ~\citeauthor{he2016deep}~\cite{he2016deep}. The final model architecture, with its pipeline depicted in Figure \ref{fig:overview}, consists of:

\begin{enumerate}
    \item an initial pre-processing stage;
    \item a sequence of many convolutional layers based on the first 49 layers of ResNet-50 and;
    \item a top classifier replacing the original 1000 fully-connected softmax layer.
\end{enumerate}

\begin{figure*}[!h]
  \centering
  \includegraphics[width=0.99\textwidth]{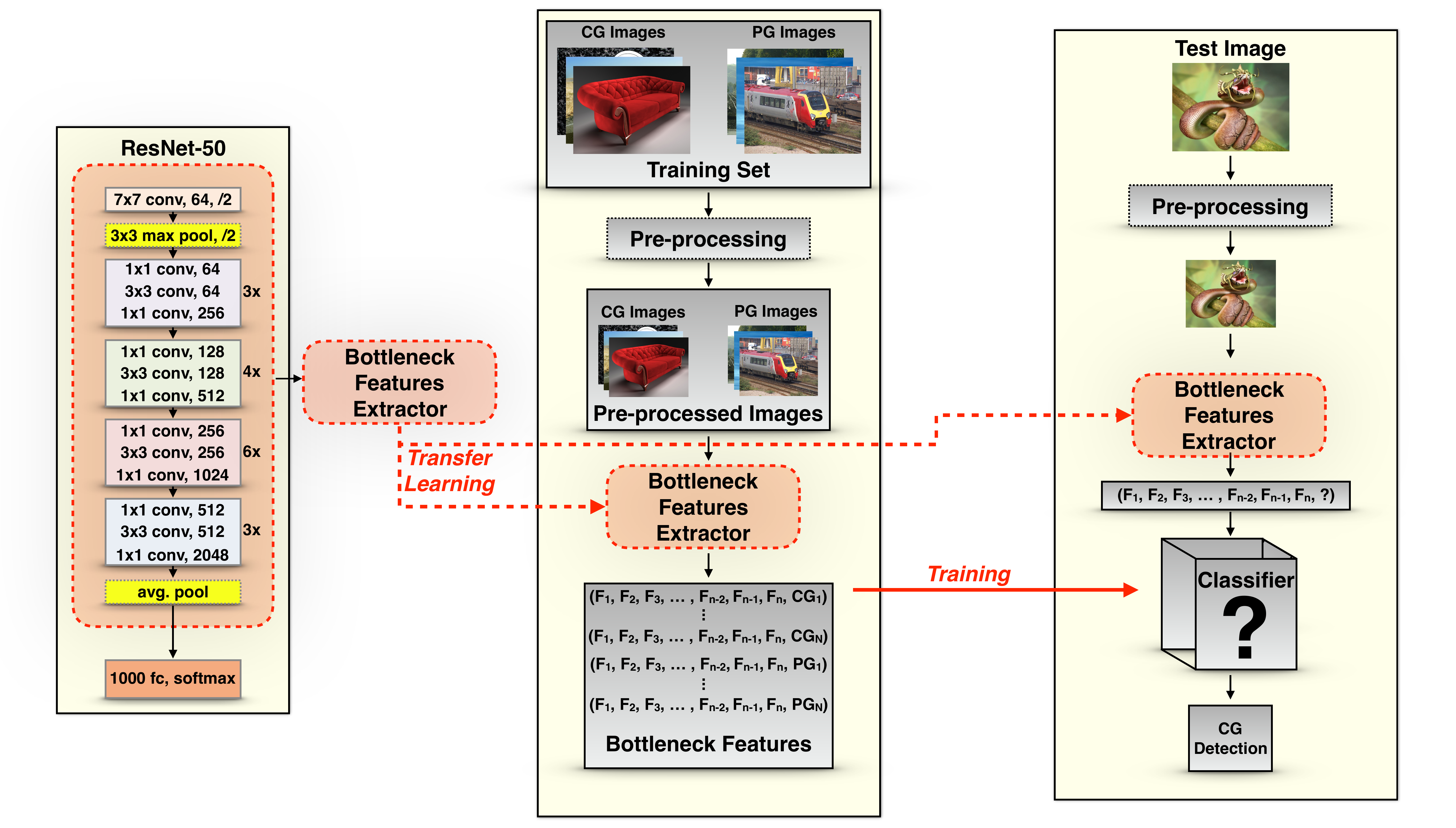}
  \caption{Overview of proposed method. Transferring ResNet-50 parameters to our model to extract bottleneck features, which are used to train different classifiers.}
 \label{fig:overview}
\end{figure*}

In a real word dataset, images can present different resolutions. However, our model requires a constant input dimensionality. Therefore, we resize the images to a fixed resolution of $224\times224$ and, for each pixel, we subtract the mean RGB value computed over the ImageNet dataset (as proposed by \citeauthor{krizhevsky2012imagenet}~\cite{krizhevsky2012imagenet}). These two operations are performed by the pre-processing layer.

After pre-processing the dataset images, we apply the transfer learning techniques explained in Section \ref{sec:transferlearning}. In our CNN model, after the pre-processing, we use the first 49 layers of ResNet-50 with their weights trained on ImageNet as a features extractor (red box named ``Bottleneck Features Extractor'' of Figure \ref{fig:overview}) to generate a set of features with the correspondent label associated. These labelled features, also called bottleneck features\footnote{Bottleneck term refers to a neural network topology where the hidden layer has significantly lower dimensionality than the input layer, assuming that such layer --- referred to as the bottleneck --- compresses the information needed for mapping the neural network input to the neural network output, increasing the system robustness to noise and overfitting. Conventionally, bottleneck features are the output generated by the bottleneck layer.} are the activation maps generated by the average pooling layer (the 49\textsuperscript{th} layer of ResNet-50), ignoring the last 1000 fully-connected softmax layer.

These bottleneck features are used to train a top classifier that will make the final prediction of CG images. This classifier has the same role as the original softmax layer at the end of ResNet-50, adapting the network to the binary problem of CG image detection. The replacement of this softmax layer, with thousands of parameters, associated with the transfer learning techniques used (no learning happens at the convolutional layers), allowed us to deploy a very deep CNN for the CG image detection problem without the requirement of millions of CG/PG labelled images, besides significantly reducing the training time. 

Once finished the training process, our final deep CNN model used for testing is made up of the pre-processing layer, the Bottleneck Features Extractor and a top classifier (network on the right of Figure \ref{fig:overview}). Different type of classifiers were trained as top predictors in order to discover which one performs best for the task of CG image detection. Section \ref{sec:classifiers} will delve into the details of each type used.

\subsection{ResNet-50}

Residual Networks (ResNet)~\cite{he2016deep} can be classified as convolutional neural networks (CNN). These CNNs, in turn, can be defined as neural networks that have at least one layer using the convolution operation~\cite{Goodfellow-et-al-2016}. Mathematically speaking, the convolution operation can be view as a weighted average operation of two functions ($x$ and $w$), where one of them ($w$) is a probability density function. More formally:

\begin{equation}
\centering
s(t) = (x * w)(t) = \int x(a) w(t-a) da
\end{equation}

In practice, due to the commutative property, the convolution operation is usually implemented as the cross-correlation function~\cite{Goodfellow-et-al-2016}. For example, assuming the function $x$ as a two-dimensional input image $I$ and the probability density function $w$ as a function $K$ (usually called \textit{kernel} in machine learning terminology), the function $S$ below is the convolution of \textit{kernel} $K$ over the image $I$:

\begin{equation}
\centering
S(i,j) = (I * K)(i,j) = \sum_{m} \sum_{n} I(i+m,j+n)K(m,n)
\end{equation}

In machine learning nomenclature, the function $S$ is usually called a \textit{feature map}. Among many interesting properties that convolutions convey, one of paramount importance is the robustness to translation in the recognition of patterns. If a kernel $K$ is specialized in recognizing circles, the convolution of this kernel over the image will identify circles no matter where they occur in the image.

Besides convolutions, CNNs usually have some non-linear activation functions (like sigmoid or rectified linear functions) and pooling layers that have the effect of turning the representation invariant to small translations in the image~\cite{Goodfellow-et-al-2016}.

Residual Networks (ResNets)~\cite{he2016deep} are deep convolutional networks where the basic idea is to skip blocks of convolutional layers by using shortcut connections to form conceptual shortcut blocks named residual blocks. The residual block can be formally described in the general form:

\begin{equation}
\centering
\mathcal{Y}_{l}(\mathbf{x},W) = f(h(\mathbf{x}) + \mathcal{F}(\mathbf{x},W))
\end{equation}
where $\mathbf{x}$ is the input of the block, $h$ is the \textit{shortcut} function (crucial to ResNet and better explained below), $\mathcal{F}$ is the mapping done by a block of one or more consecutive convolutional layers (being skipped), $W$ is the weights of these convolutional layers, $f$ is a rectified linear unit (ReLU) and $\mathcal{Y}_{l}$ is the mapping that describes the residual block $l$ as a function of the input $\mathbf{x}$ and the weights $W$. The shortcut function is usually the identity function (or a very simple convolutional layer when a dimensional compatibility is needed) and its purpose is to speed the learning of the mapping $\mathcal{F}$ as a perturbation of the input $\mathbf{x}$, starting this learning from a point near the identity function. This is faster to learn than from a random point near the zero-mapping (as done by previous architectures) and is one of the key contributions of Residual Networks.


In ResNet-50 architecture, the basic residual blocks, also called bottleneck blocks, are composed of a sequence of three convolutional layers with filters of size $1\times1$, $3\times3$ and $1\times1$ respectively. The down-sampling is performed directly by convolutional layers that have a stride of 2 and batch normalization \cite{ioffe2015batch} is performed right after each convolution and before ReLU activation.


The identity shortcuts can be directly used when the input and output feature maps are of the same dimensions. When the dimensions change, two options are considered: (i) The shortcut still performs identity mapping, with extra zero entries padded for increasing filter dimensions (depth). This option introduces no extra parameter; (ii) A $1\times1$ convolution layer is used to match the dimensions. This is called  \textit{projection} shortcut. For both options, when the shortcuts go across feature maps of two sizes, they are performed with a stride of 2. Figure~\ref{fig:bottleneckblocks} tries to clarify this set-up.

\begin{figure*}[!h]
  \centering
  \includegraphics[width=0.99\textwidth]{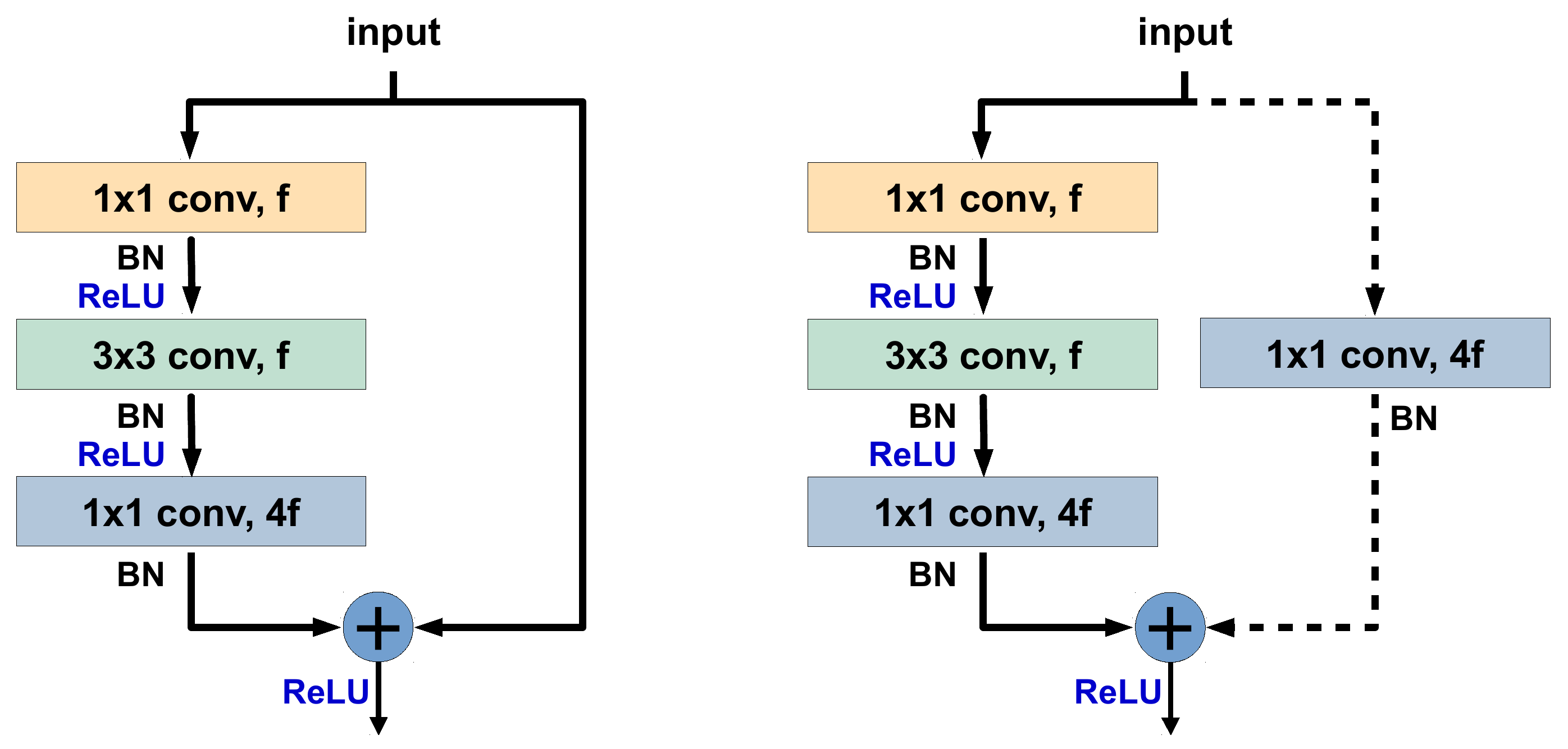}
  \caption{Bottleneck Blocks for ResNet-50 (left: identity shortcut; right: projection shortcut).}
 \label{fig:bottleneckblocks}
\end{figure*}

The network ends with a global average pooling layer and a 1000-way fully-connected layer with softmax activation. The total number of weighted layers is 50.

\subsection{Transfer Learning}
\label{sec:transferlearning}

Deep learning techniques usually require a very large number of samples and demands a heavy computing effort. In this scenario, transfer learning ~\cite{yosinski2014transferable} has gained a lot of attention. It represents the possibility to transfer the knowledge learned from one problem to another problem. For neural networks, the transfer learning process consists, in practice, in transferring the parameters of a (source) neural network that was previously trained for a particular dataset for a specific task to another (target) network with a different dataset to solve a different problem.

A typical transfer learning procedure implies in using an already trained base network and copying all the parameters from the first $n$ layers to the first $n$ layers of the target network. Then, a supervised training is performed only on the remaining layers that were not copied. Since the task is different and probably so is  number of classes, the last layer has to be modified to contain the same amount of neuron as the number of classes or it can be replaced by another classifier.
During this training, the copied layers can be left fixed, or it can fine-tuned by allowing the backpropagation process into the copies parameters. Usually fine-tuning  improves the accuracy, however, when the number of parameters is big and the numbers of samples is small, this may result in overfitting and fine-tuning should not be used.

In traditional supervised learning, the common sense was settled that the training should always be performed specifically for a given task and dataset and the transfer learning approach could sound senseless. However, in general, deep neural networks present a particular characteristic which is to learn features in the first layers that are more general and not specific to that particular dataset and problem. That makes that knowledge reusable for other problems and dataset. On the other hand, the knowledge learned on the last layers of the neural network are more specific for that task.

Transfer learning is very handy because it avoids the  task of training the network. Since the architecture is very deep, training it represents an enormous computational efforts requiring expensive high processing computers, usually using multiple high-end GPU units. Another problem in training deep networks from scratch is that it required a vary large number of samples, otherwise is will overfit the data. Transfer learning significantly mitigates this two problems, enabling the use of deep learning even when the target dataset is small or when there is limited computing resources. Recent studies have taken advantage of this fact to obtain state-of-the-art results~\cite{donahue2014decaf}\cite{zeiler2014visualizing}\cite{sermanet2013overfeat}, which evidences the generality of the features learned in the first layers of the network.

In this work, we apply transfer learning technique by using the same parameters of the ResNet-50 network trained for the ImageNet 2012 competition~\cite{russakovsky2015imagenet} which provided a 1.28 million images dataset from 1000 classes. The first 49 layers of the network were copied to a new ResNet-50 network and we removed the last layer, replacing it by another classifier. We have not used fine-tuning during the training due to the relatively small dataset.

\subsection{Top Classifier}
\label{sec:classifiers}

In the proposed method, the last layer of the ResNet-50 is replaced by another classifier and we have evaluated four different classifier for this task.

\subsubsection{Softmax}

The softmax function\cite{Bishop:2006:PRM:1162264} is widely used in deep learning architectures and consists in the generalization of the binary logistic regression to multiple classes. 

This function is particularly interesting because it provides an intuitive output with probabilist interpretation. The outcome of the function is a vector containing the probability for each class.

Typically, the softmax is used for classification as the activation function on the last fully-connected layer of CNNs, which is the case in the ResNet-50. The function transforms a vector $\mathbf{z}$ with dimension K of real numbers ${z_k}$, to another vector $\sigma(\mathbf{z})$ of same dimensions with the values ranging from 0 to 1. The sum of the output vector adds up to 1, therefore it can be interpreted as the probabilities for each class. The formula is given by:

\begin{equation}
\centering
\sigma(z)_{j}=\frac{\mathrm{e}^{z_j}}{\sum_{k=1}^K\mathrm{e}^{z_k}}
\end{equation}

For training, we used the categorical cross-entropy loss function, which is given by:

\begin{equation}
\centering
H(p,q)=-\sum_xp(x) . log(q(x))
\end{equation}

\subsubsection{k-Nearest Neighbor (kNN)}

The k-Nearest Neighbor (kNN) \cite{Bishop:2006:PRM:1162264} classifier is one of the simplest and most popular supervised classifier. It consists in classifying a sample based on the k nearest samples from the training set. Usually, the euclidean distance function is used, but other distance functions may be chosen. The sample is then classified by the majority voting or other similar function among the k nearest samples. The advantage of kNN is that it is very simple, does not require an explicit training step and yet it is very effective for many applications, specially when the training set is large. The main drawbacks of this method are that (i) the method requires the distance computation to all samples from the training dataset, which makes it computationally heavy; and (ii) the fact that it also requires a lot of memory since the whole dataset has to be loaded for comparison.

\subsubsection{XGBoost}

Extreme Gradient Boosting (XGBoost)~\cite{Chen_and_Guestrin_2016} is a recent work that has been gaining a lot of attention for impressive results in machine learning challenges like KDD Cup and Kaggle competitions.

XGboost consists of an open-source package that implements gradient tree boosting algorithm with the focus on being highly effective and scalable. It includes novel optimized algorithms related to: more efficient parallelization; a novel sparse-aware tree learning algorithm; out-of-core computation, cache-aware access; distributed weighted quantile sketch method; among other improvements. All these improvements allow the system to perform more than ten times faster than other tree boosting solutions. 

The library was written in C++, however binding for other languages like python, R and Java are available.

\subsubsection{SVM}

Support Vector Machines (SVM)\cite{Bishop:2006:PRM:1162264} is one of the most popular supervised classifier and basically operates by finding the optimum hyperplane that best separates two classes. Originally designed for binary classification, it can be extended for multi-class problems by reducing one multi-class task to multiple binary classification problems using techniques known as one-versus-one or one-versus-all, among other methods. 

Although the original SVM is a linear classifier, it can be applied for non-linear problems by using kernel functions which nonlinearly maps the feature vector to a new space. 

In the present work, we evaluate both the linear SVM and  the SVM with a Radial Basis Function (RBF) kernel, for the top classifier of the architecture.

\section{Experiments and Results}
\label{sec:results}

To validate the proposed approach, we have performed different rounds of experiments which will be detailed in the next sections.

\subsection{DSTok Dataset}

The first dataset in which the proposed method has been tested is a public dataset proposed by \citeauthor{Tokuda20131276}~\cite{Tokuda20131276}. It comprises 4,850 CG images and 4,850 PG images, depicting different kinds of scenarios as people, outdoor, objects, cars, animals, and others. The entire set of images has been collected from Internet and compressed in JPEG format, presenting images with sizes from 12 KB to 1.8 MB. Images in the dataset present different resolutions and, differently from \citeauthor{Tokuda20131276}, our proposed method works with the entire image (without cropping). Figure~\ref{fig:dstok} depicts some examples of images in DSTok dataset.

\begin{figure}[h]
  \centering
  \subfloat[CG]{\includegraphics[height=4.3cm]{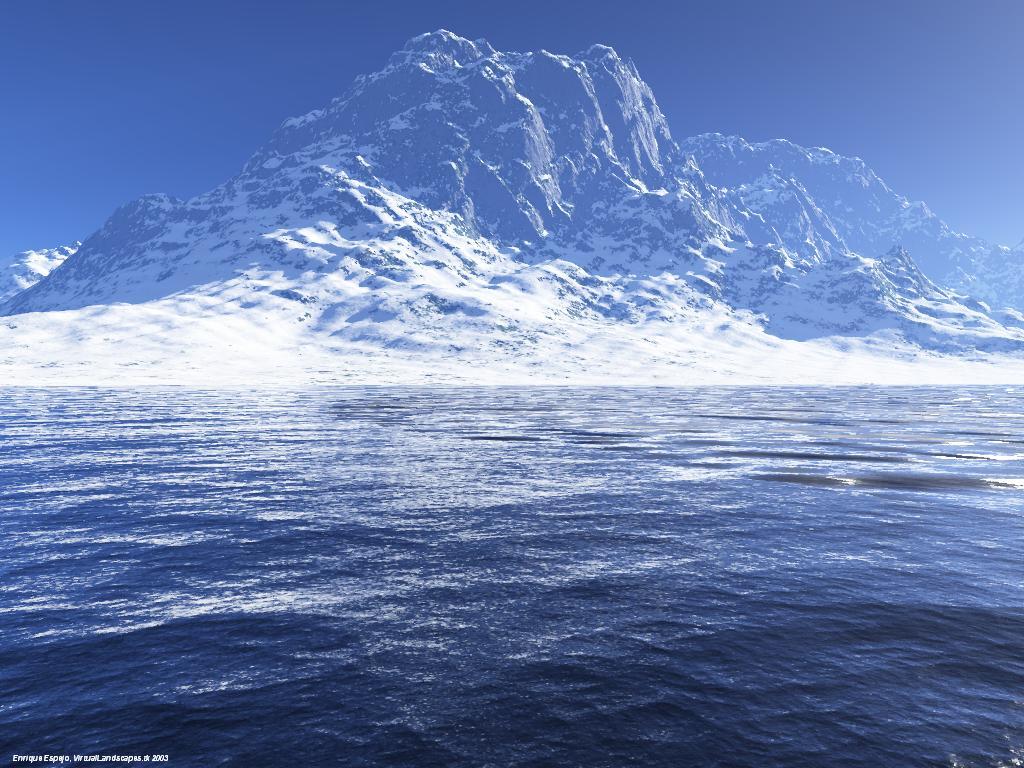}} \hspace*{0.05cm}
  \subfloat[CG]{\includegraphics[height=4.3cm]{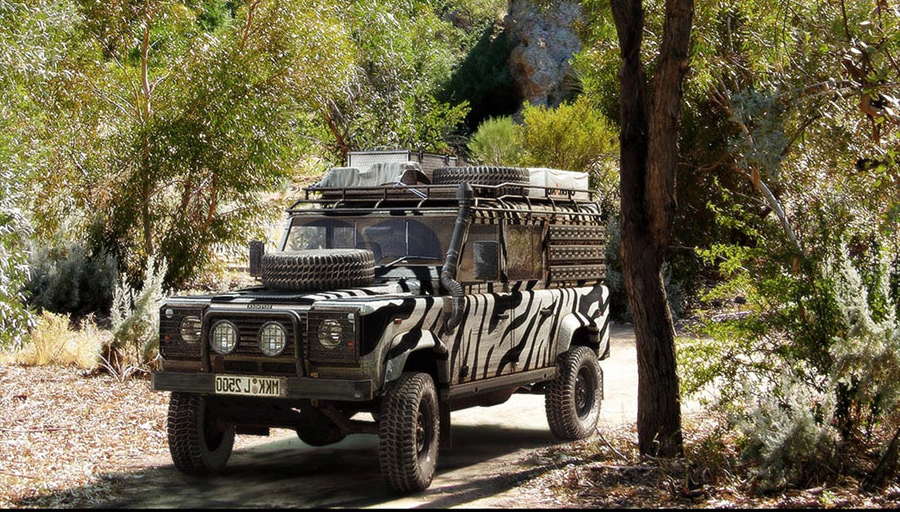}}\\
  \subfloat[PG]{\includegraphics[height=4.3cm]{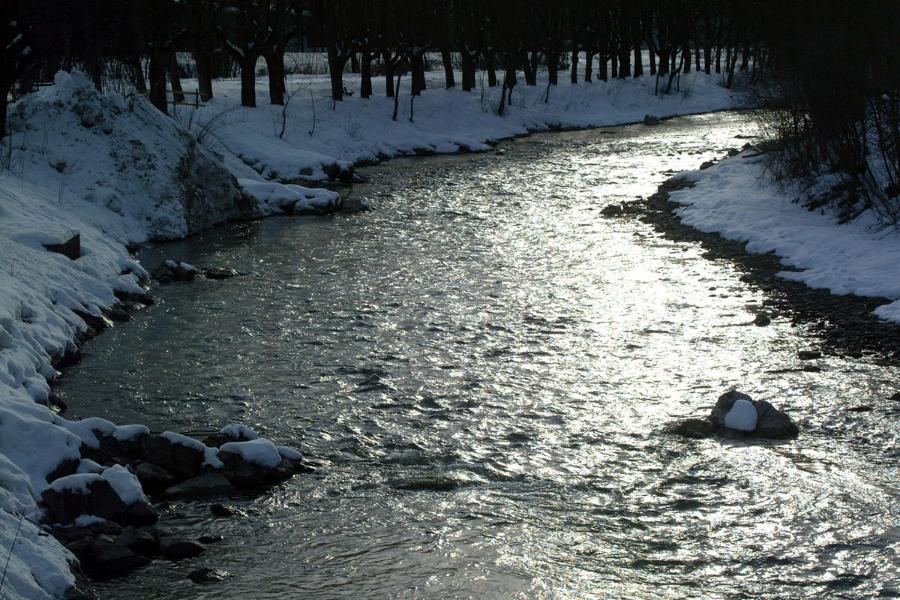}} \hspace*{0.05cm}
  \subfloat[PG]{\includegraphics[height=4.3cm]{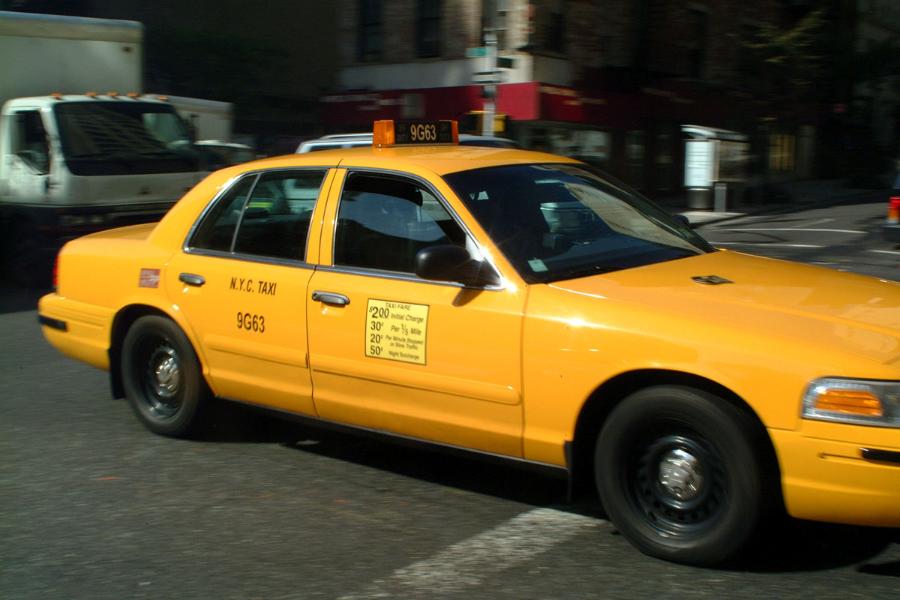}}
  \caption{Examples of images in DSTok dataset.}
 \label{fig:dstok}
\end{figure}

\subsection{DSTokExt Dataset}
\label{sec:dstokext}

The second dataset used to validate the method is an extension of \citeauthor{Tokuda20131276}~\cite{Tokuda20131276} dataset. It comprises 8,394 CG images and 8,002 PG images, also depicting different kinds of scenario. In the same way as \citeauthor{Tokuda20131276}~\cite{Tokuda20131276}, all the images have been collected from Internet and compressed in JPEG format, presenting images with sizes from 12 KB to 1.8 MB. Images in the dataset present different resolutions. Figure~\ref{fig:dstokext} depicts some examples of images in DSTok dataset.

\begin{figure}[h]
  \centering
  \subfloat[CG]{\includegraphics[height=4.3cm]{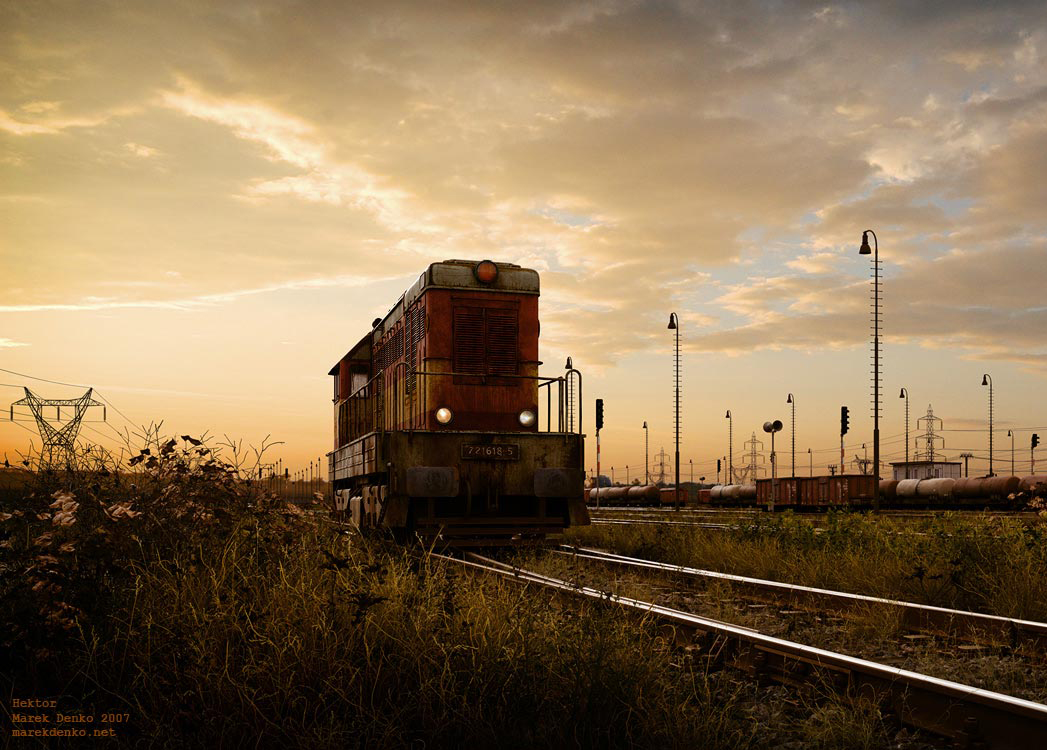}} \hspace*{0.05cm}
  \subfloat[CG]{\includegraphics[height=4.3cm]{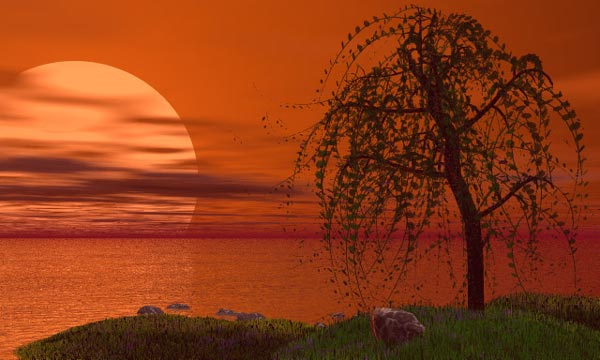}}\\
  \subfloat[PG]{\includegraphics[height=4.3cm]{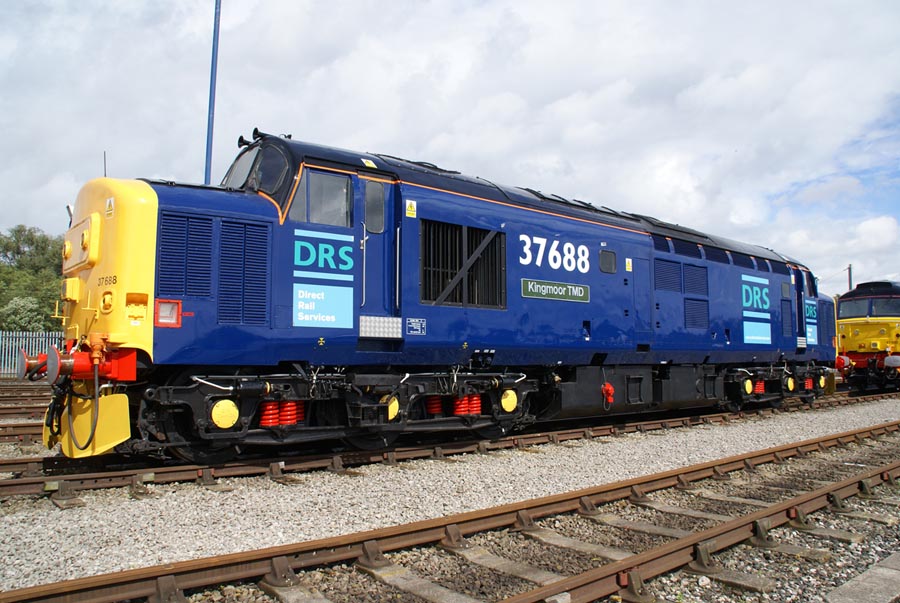}} \hspace*{0.05cm}
  \subfloat[PG]{\includegraphics[height=4.3cm]{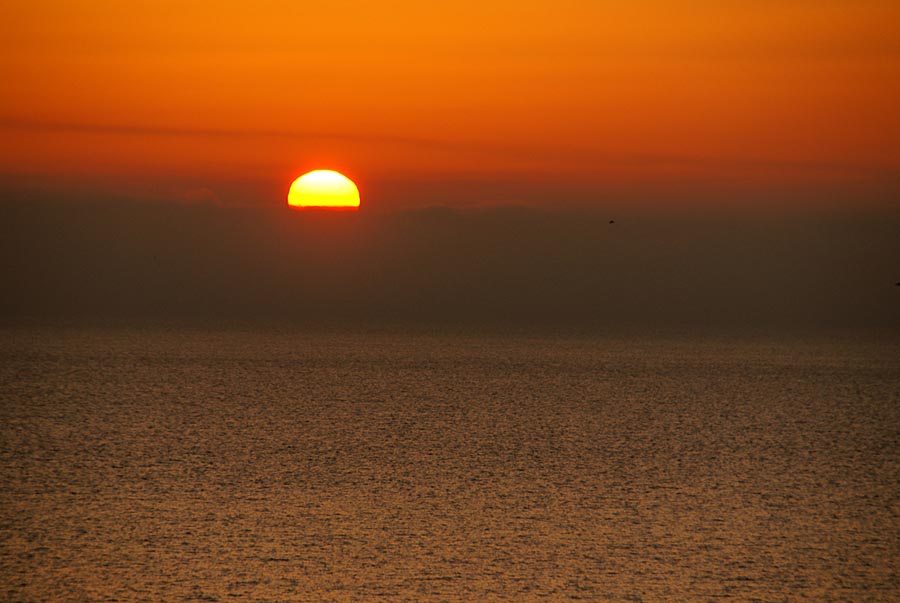}}
  \caption{Examples of images in DSTokExt dataset.}
 \label{fig:dstokext}
\end{figure}

\subsection{Validation Protocol}

In order to compare the results achieved by the proposed method with the results reported by \etal{Tokuda}~\cite{Tokuda20131276}, we perform the same five fold cross-validation protocol, reporting the accuracy by fold and the average accuracy for each round of experiments.

\subsection{Implementation Details}

The proposed methods have been implemented using Python 3.5, Keras 2.0.3\footnote{\url{https://keras.io}}, and TensorFlow 1.0.1\footnote{\url{https://www.tensorflow.org}}. All performed tests have been executed in a machine with an Intel(R) Xeon(R) CPU E5-2620 2.00GHz processor with 96GB of RAM and two Nvidia Titan Xp GPUs.

\subsection{Round \#1: ResNet-50 trained from Glorot uniform initialization over DSTok}
\label{sec:round1}

In the first round of experiments we classify samples from DSTok using a deep CNN architecture similar to the original ResNet-50. Since we have only 2 classes (CG and PG), we have adapted the ResNet-50 architecture to the CG detection task replacing its last 1,000 fully-connected softmax layer by a 2 fully-connected softmax layer. The weights of the network have been initialized using Glorot uniform approach~\cite{glorot2010understanding} and the bias terms were initialized to zero. All layers of the model have been trained for 200 epochs with categorical cross-entropy cost function and Adam optimizer ($lr=0.001$, $beta_1=0.9$, $beta_2=0.999$, $epsilon=1e-08$ and $decay=0.0$).

Figures \ref{fig:ResNet50-Fromscratch-Loss} and \ref{fig:ResNet50-Fromscratch-Accuracy} present, respectively, the average loss and accuracy of ResNet-50 trained from Glorot uniform initialization. Solid lines represent the average performance in the training (red) and testing (green) set while the shadows represent the standard deviation in the 5-fold cross-validation. 

\begin{figure}[h]
  \centering
  \includegraphics[width=.99\textwidth]{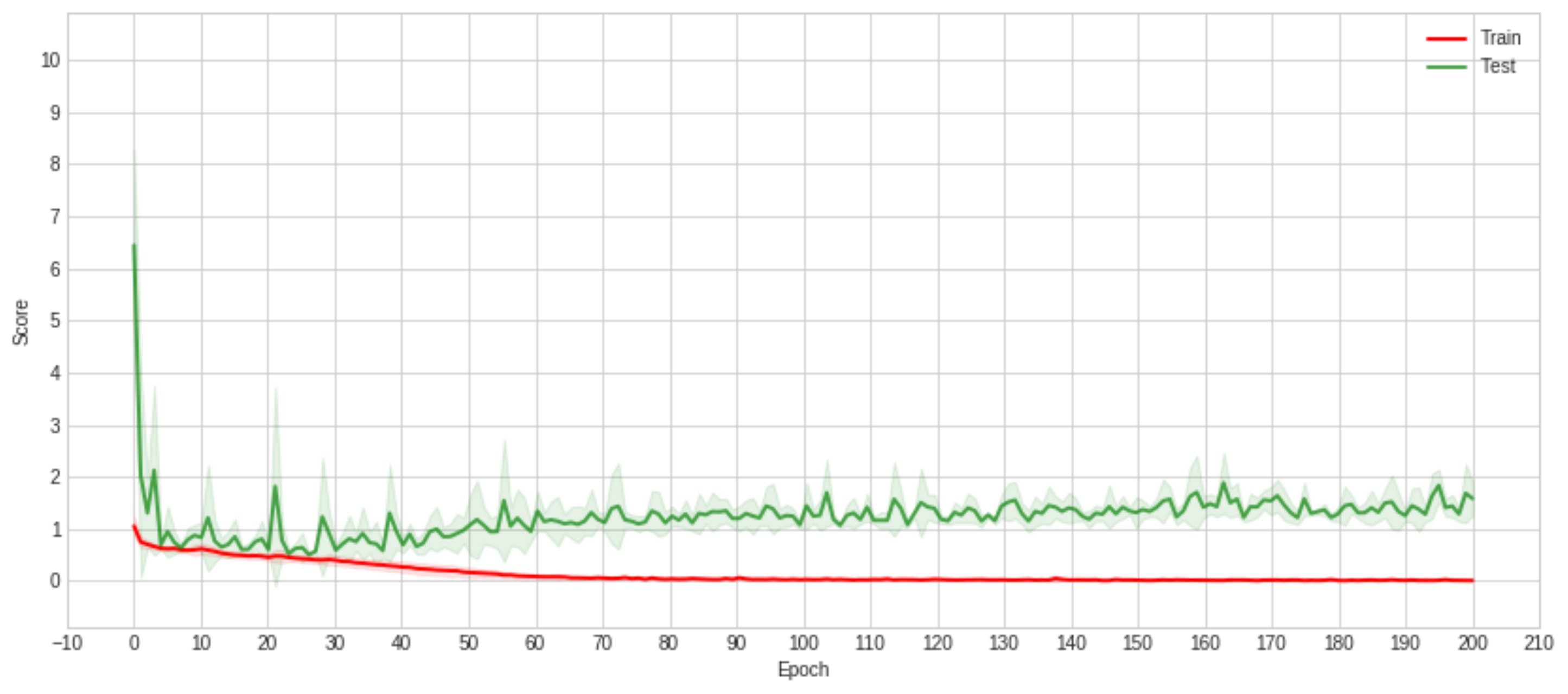}
  \caption{Train and test average loss of ResNet-50 trained from Glorot uniform initialization.}
  \label{fig:ResNet50-Fromscratch-Loss}
\end{figure}

\begin{figure}[h]
  \centering
  \includegraphics[width=.99\textwidth]{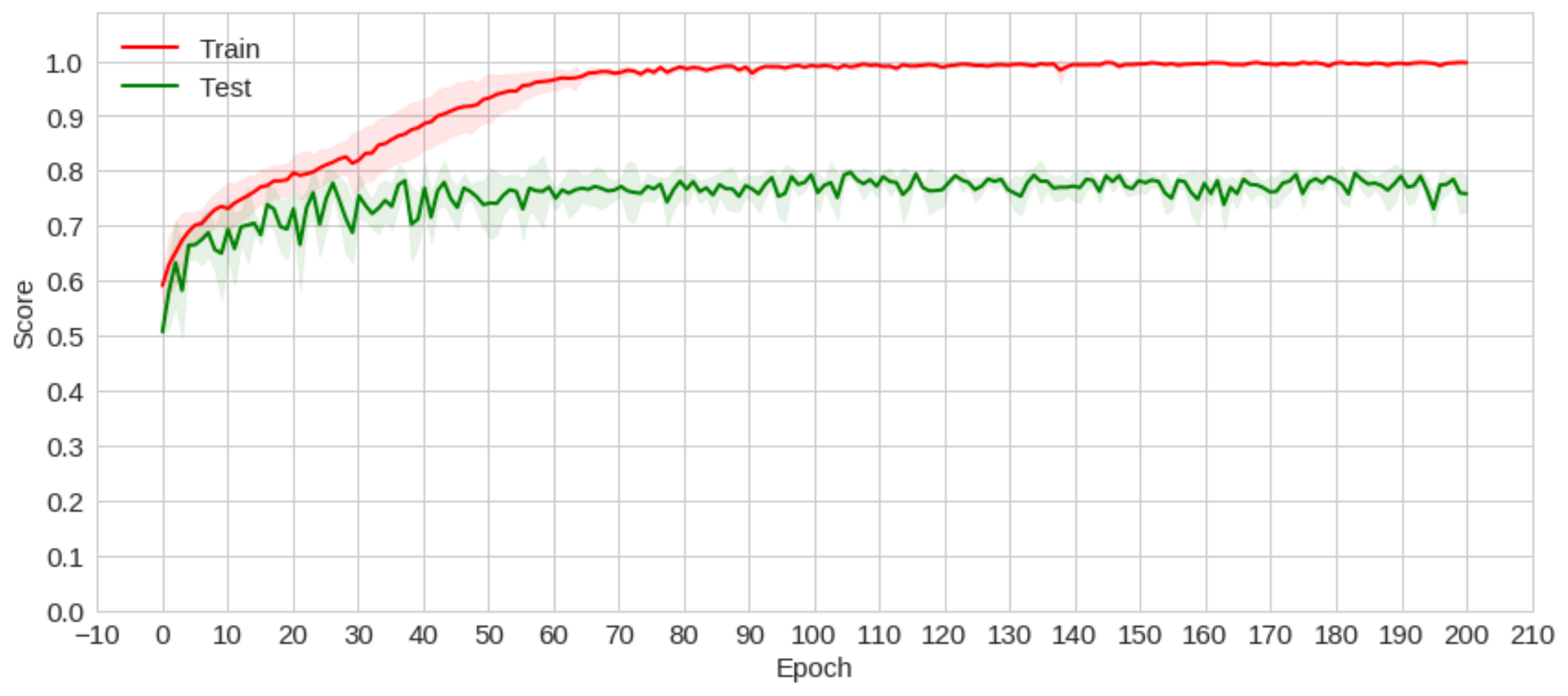}
  \caption{Train and test average accuracy of ResNet-50 trained from Glorot uniform initialization.}
  \label{fig:ResNet50-Fromscratch-Accuracy}
\end{figure}

The results by fold fold are presented in Table~\ref{table:ResNet50-Fromscratch-Accuracy}. This table shows that the 75.85\% average accuracy was achieved with a training time around 17,369.90 seconds after 200 training epochs. 

\begin{table}[h]
\centering
\scriptsize
\caption{Accuracy by fold of ResNet-50 trained from Glorot uniform initialization.}
\label{table:ResNet50-Fromscratch-Accuracy}
\begin{tabular}{|p{1.2cm}|p{2cm}|p{2cm}|}
\hline
\rowcolor[HTML]{9B9B9B}
\textbf{Fold}    & \textbf{Accuracy} & \textbf{Time (s)}    \\ \hline
\rowcolor[HTML]{FFFFFF}
0       & 0.79   & 17,387.85   \\
\rowcolor[HTML]{EFEFEF}
1       & 0.72   & 17,352.25   \\
\rowcolor[HTML]{FFFFFF}
2       & 0.80   & 17,357.33   \\
\rowcolor[HTML]{EFEFEF}
3       & 0.74   & 17,353.05   \\
\rowcolor[HTML]{FFFFFF}
4       & 0.74   & 17,399.03   \\ 
\rowcolor[HTML]{EFEFEF}
{\color[HTML]{3531FF} \textbf{Average}}& {\color[HTML]{3531FF} \textbf{0.76}}   & {\color[HTML]{3531FF} \textbf{17,369.90}}   \\ \hline
\end{tabular}
\end{table}

\subsection{Round \#2: ResNet-50 fine-tuned from ImageNet initialization over DSTok}
\label{sec:round2}

In the second round of experiments, we evaluate the impact of transfer learning as a strategy to initialize the weights of the convolutional layers in the proposed model. We transfer the weights of ResNet-50 convolutional layers pre-trained on ImageNet dataset to our deep CNN model, replacing the last 1000 fully-connected softmax layer by a 2 fully-connected softmax layer.

In the first round of experiments, all network parameters (including the last layer) have been initialized using Glorot uniform approach and the bias terms were initialized to zero. At this round of experiments, we use ImageNet parameters as initial weights, except in the last layer where, again, we apply Glorot uniform initialization. Then, all layers have been trained for 200 epochs with categorical cross-entropy cost function and Adam optimizer ($lr=0.001$, $beta_1=0.9$, $beta_2=0.999$, $epsilon=1e-08$ and $decay=0.0$).

Figures \ref{fig:ResNet50-Finetune-Loss} and \ref{fig:ResNet50-Finetune-Accuracy} present, respectively, the average loss and accuracy of ResNet-50 fine-tuned from ImageNet initialization. Solid lines represent the average performance in the training (red) and testing (green) set while the shadows represent the standard deviation in the 5-fold cross-validation.

\begin{figure}[h]
  \centering
  \includegraphics[width=.99\textwidth]{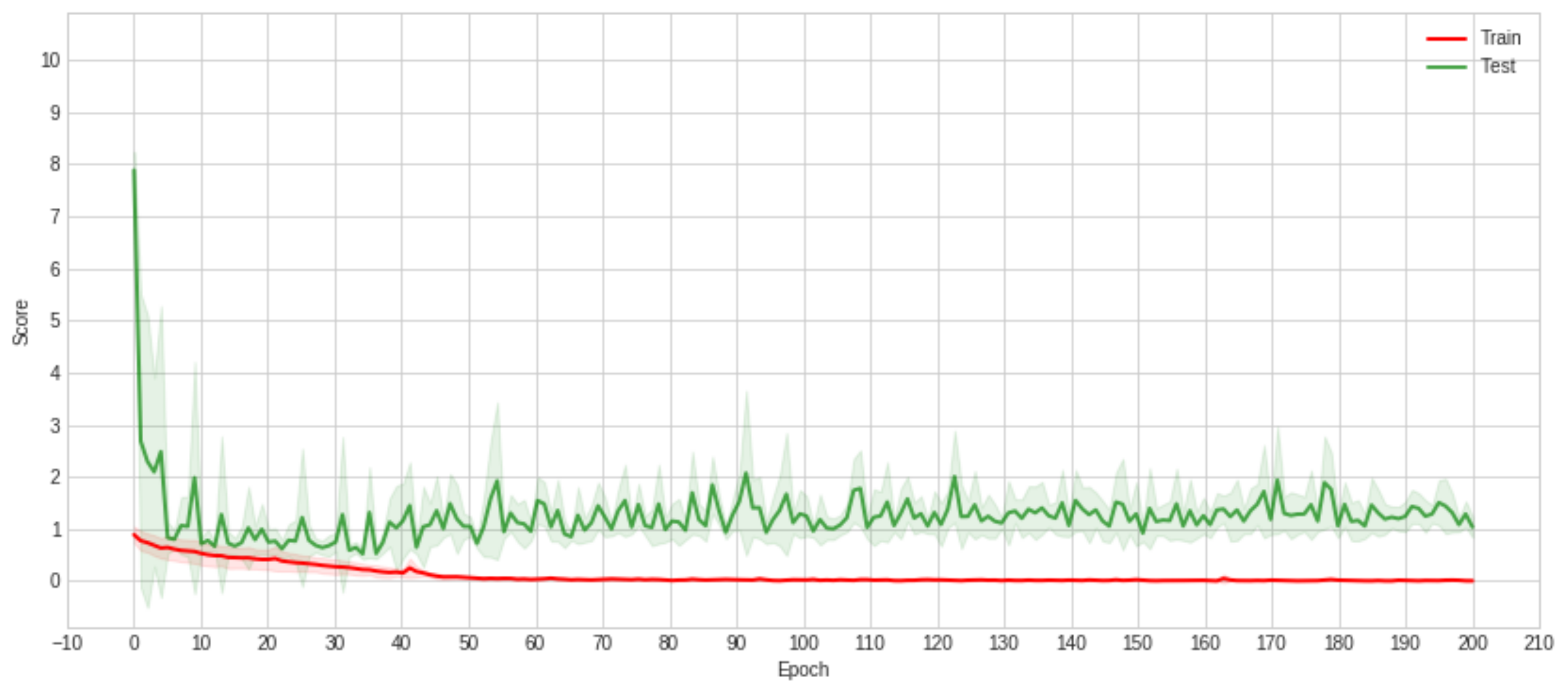}
  \caption{Train and test average loss of ResNet-50 fine-tuned from ImageNet initialization.}
  \label{fig:ResNet50-Finetune-Loss}
\end{figure}

\begin{figure}[h]
  \centering
  \includegraphics[width=.99\textwidth]{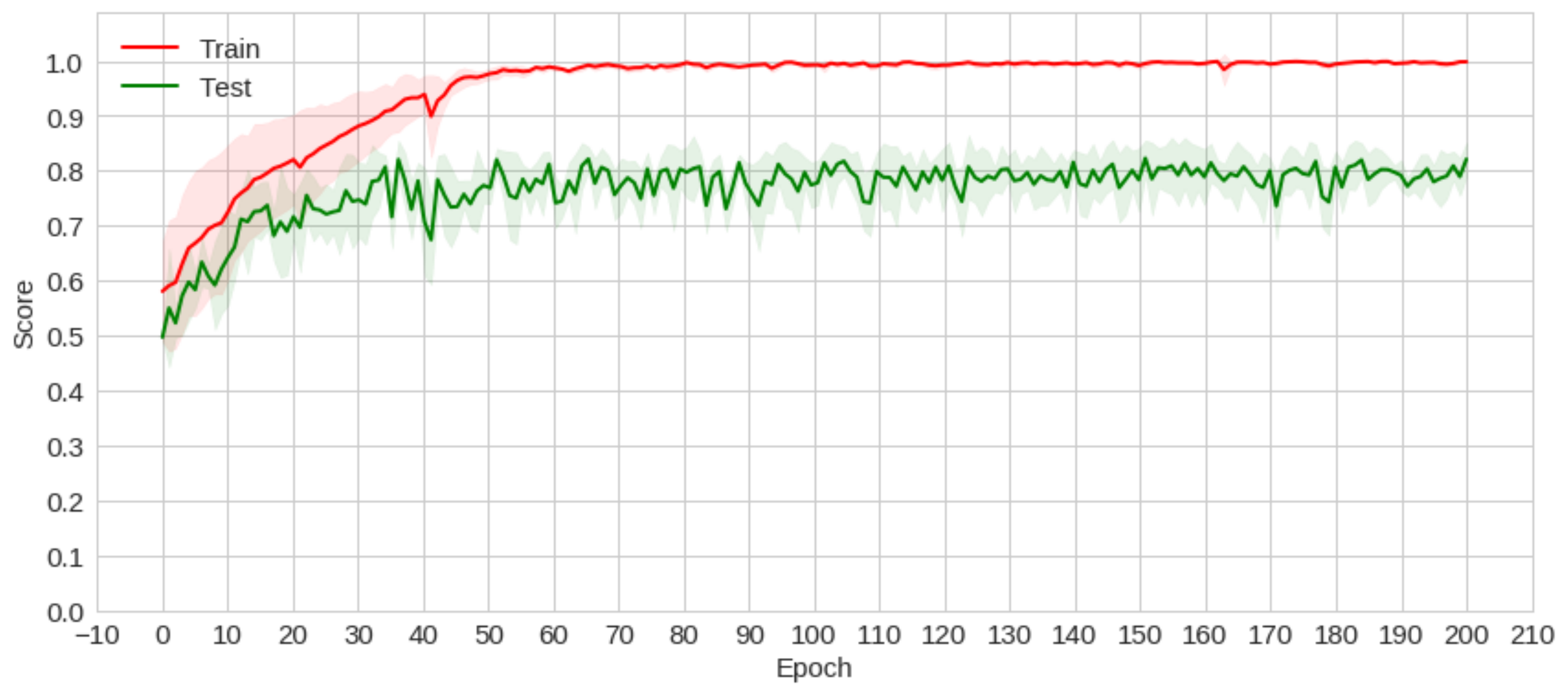}
  \caption{Train and test average accuracy of ResNet-50 fine-tuned from ImageNet initialization.}
  \label{fig:ResNet50-Finetune-Accuracy}
\end{figure}

The results by fold are presented in Table~\ref{table:ResNet50-Finetune-Accuracy}. This table shows that the 82.12\% average accuracy was achieved with a training time around 16,889.14 seconds after 200 training epochs. 

\begin{table}[h]
\centering
\scriptsize
\caption{Accuracy by fold of ResNet-50 fine-tuned from ImageNet initialization.}
\label{table:ResNet50-Finetune-Accuracy}
\begin{tabular}{|p{1.2cm}|p{2cm}|p{2cm}|}
\hline
\rowcolor[HTML]{9B9B9B}
\textbf{Fold}    & \textbf{Accuracy} & \textbf{Time (s)}    \\ \hline
\rowcolor[HTML]{FFFFFF}
0       & 0.86   & 16,888.32   \\
\rowcolor[HTML]{EFEFEF}
1       & 0.82   & 16,895.00   \\
\rowcolor[HTML]{FFFFFF}
2       & 0.85   & 16,897.95   \\
\rowcolor[HTML]{EFEFEF}
3       & 0.77   & 16,879.80   \\
\rowcolor[HTML]{FFFFFF}
4       & 0.80   & 16,884.62   \\
\rowcolor[HTML]{EFEFEF}
{\color[HTML]{3531FF} \textbf{Average}} & {\color[HTML]{3531FF} \textbf{0.82}}   & {\color[HTML]{3531FF} \textbf{16,889.14}}   \\ \hline
\end{tabular}
\end{table}

Given the improvement of 7\% over the average accuracy obtained in the experiments of Round \#1, we can conclude that the knowledge transferred from ImageNet dataset to CG image detection problem produced good results.

\subsection{Round \#3: ResNet-50 fine-tuned from ImageNet initialization and pre-trained softmax layer over DSTok}
\label{sec:round3}

In Round \#2 of experiments, we showed that the transfer learning in fact helps to improve the model accuracy of CG detection task. However, the random initialization of the last fully-connected softmax layer could cause an undesired drawback backpropagating the error from the last layer to the ImageNet transferred weights during the fine-tune step along the entire network, degrading the model accuracy. 

Therefore, in this round of experiments, we pre-train the last fully-connected softmax layer before the fine-tuning step along the entire network. To perform this, we initialize the convolutional layers with ImageNet weights and freeze them. Then, the weights of the last layer are initialized using Glorot uniform approach, the bias terms are initialized to zero and the network is pre-trained for 200 epochs with categorical cross-entropy cost function and Adam optimizer ($lr=0.001$, $beta_1=0.9$, $beta_2=0.999$, $epsilon=1e-08$ and $decay=0.0$). This procedure results in the training of the softmax layer, only. After this step, we unfreeze convolutional layers, training all layers of the model for 200 epochs with categorical cross-entropy cost function and SGD optimizer ($lr=0.0001$, $momentum=0.9$, $decay=0.0$ and $nesterov=False$), using a smaller learning rate to train the network. Since we expect the pre-trained weights to be quite good already as compared to randomly initialized weights, we do not want to distort them too quickly and too much.

Figures \ref{fig:ResNet50-Pretrain-Finetune-Loss} and \ref{fig:ResNet50-Pretrain-Finetune-Accuracy} present, respectively, the average loss and accuracy of ResNet-50 fine-tuned from ImageNet initialization and pre-trained softmax layer. Solid lines represent the average performance in the training (red) and testing (green) set while the shadows represent the standard deviation in the 5-fold cross-validation. 

\begin{figure}[h]
  \centering
  \includegraphics[width=.99\textwidth]{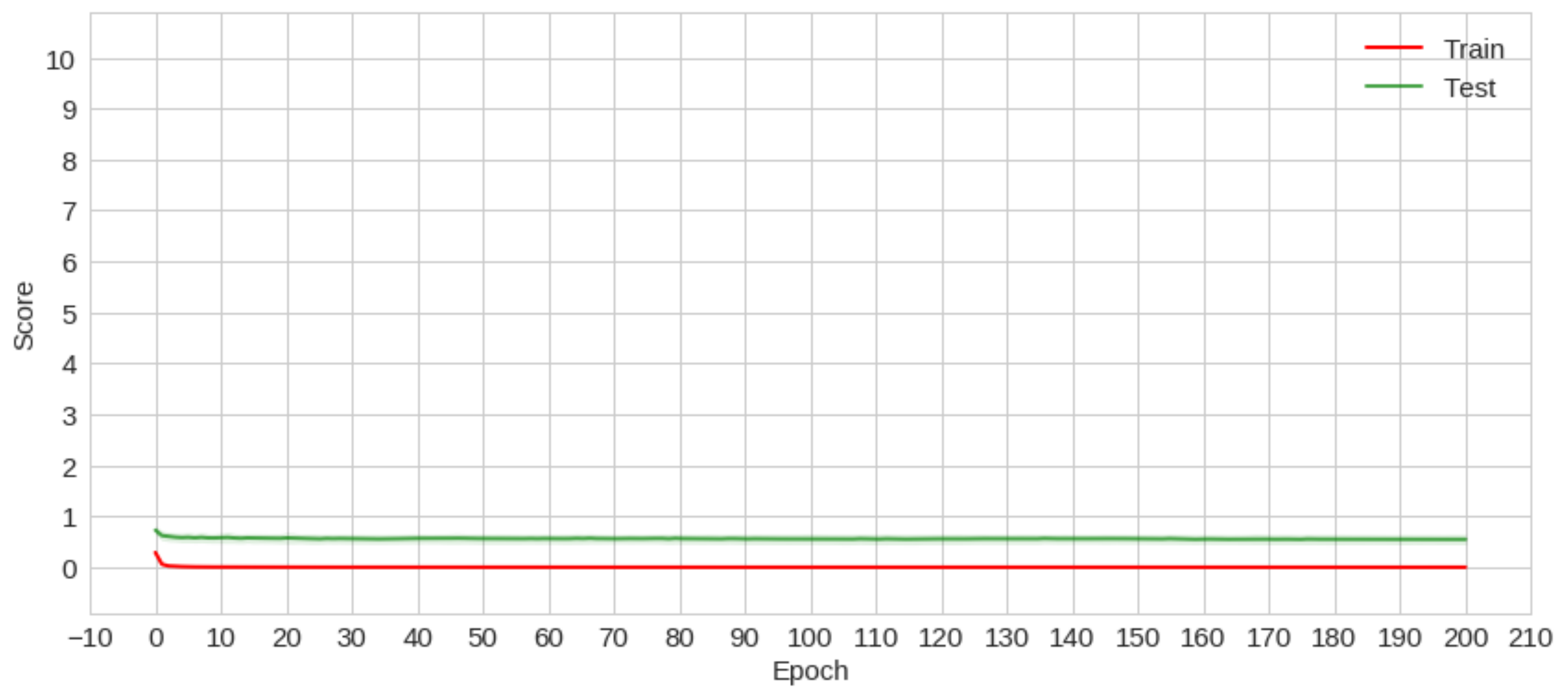}
  \caption{Train and test average loss of ResNet-50 fine-tuned from ImageNet initialization and pre-trained softmax layer.}
  \label{fig:ResNet50-Pretrain-Finetune-Loss}
\end{figure}

\begin{figure}[h]
  \centering
  \includegraphics[width=.99\textwidth]{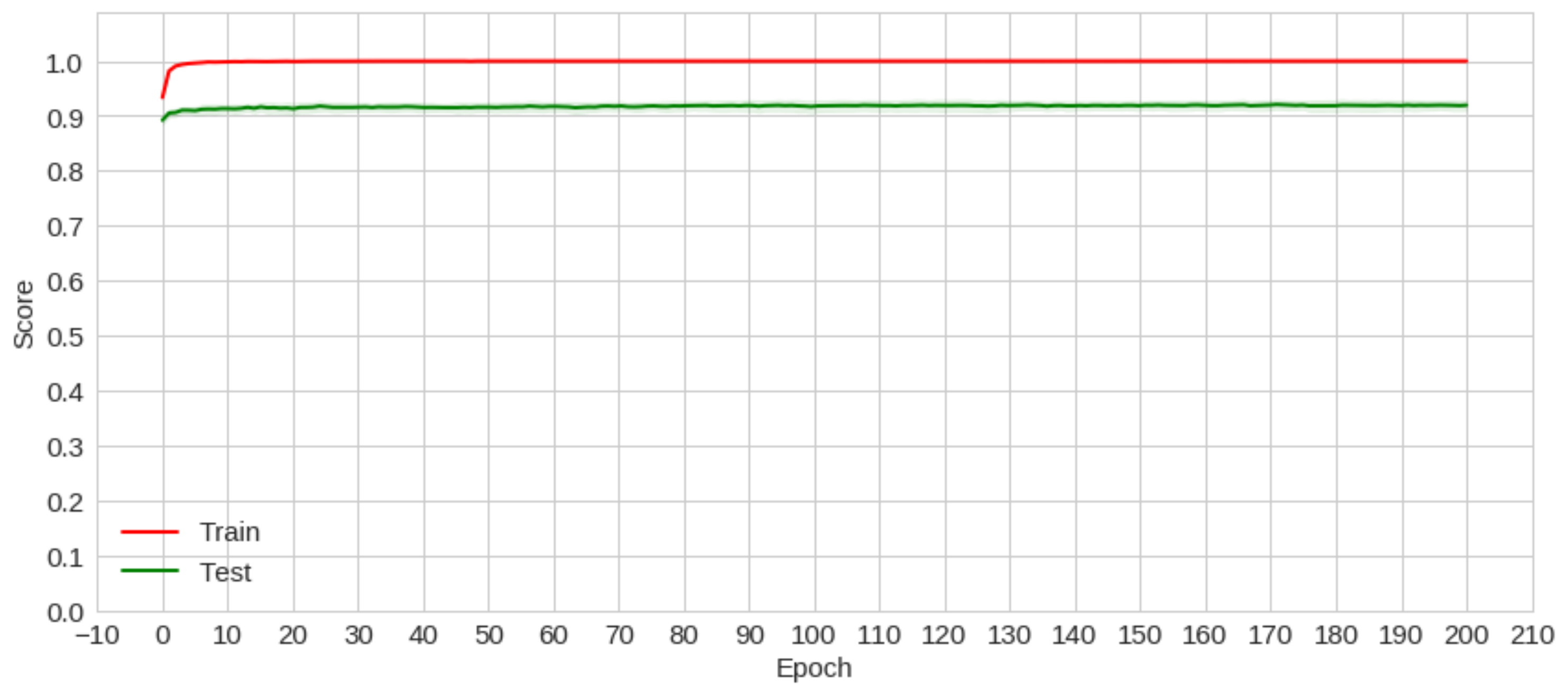}
  \caption{Train and test average accuracy of ResNet-50 fine-tuned from ImageNet initialization and pre-trained softmax layer.}
  \label{fig:ResNet50-Pretrain-Finetune-Accuracy}
\end{figure}

The results by fold are presented in Table~\ref{table:ResNet50-Pretrain-Finetune-Accuracy}. As can be seen in the table, the 2 fully-connected softmax layer pre-trained with bottleneck features (obtained freezing ImageNet weights in convolutional layers) achieved an average accuracy of 89.90\% after 200 pre-training epochs and the model fine-tuned after this pre-training step achieved an average accuracy of 91.96\% with a training time around 16,634.51 seconds after 200 training epochs.

\begin{table}[h]
\centering
\scriptsize
\caption{Accuracy by fold of 2 fully-connected softmax layer trained with bottleneck features and ResNet-50 fine-tuned from ImageNet initialization.}
\label{table:ResNet50-Pretrain-Finetune-Accuracy}
\begin{tabular}{|p{1.2cm}|p{1.8cm}|p{1.7cm}|p{2cm}|}
\hline
\rowcolor[HTML]{9B9B9B}
\textbf{Fold}    & \textbf{Pre-train Accuracy}  & \textbf{Model\qquad Accuracy} & \textbf{Time (s)}    \\ \hline
0       & 0.91              & 0.93         & 16,716.76   \\
\rowcolor[HTML]{EFEFEF}
1       & 0.89              & 0.90         & 16,618.89   \\

2       & 0.91              & 0.92         & 16,596.87   \\
\rowcolor[HTML]{EFEFEF}
3       & 0.89              & 0.91        & 16,620.85   \\

4       & 0.90              & 0.93         & 16,619.18   \\
\rowcolor[HTML]{EFEFEF}
{\color[HTML]{3531FF} \textbf{Average}} & {\color[HTML]{3531FF} \textbf{0.90}} & {\color[HTML]{3531FF} \textbf{0.92}} & {\color[HTML]{3531FF} \textbf{16,634.51}}   \\ \hline
\end{tabular}
\end{table}

Those results confirm our hypothesis that, despite the disparity between object detection and CG detection tasks, ResNet-50 comprehensively trained on the large-scale well-annotated ImageNet may still be transferred to make CG detection task more effective. Furthermore, it is important to highlight that ResNet-50 bottleneck features provide a very discriminative image descriptor for CG detection problem.

\subsection{Round \#4: ResNet-50 bottleneck features with Shallow Classifiers over DSTok}

Based on the promising results obtained with the knowledge transfer of ResNet-50 convolutional layers pre-trained on ImageNet dataset, in this round of experiments we evaluate the performance of transfer learning combined with shallow classifiers.

Therefore, we replace the last fully-connected softmax layer of ResNet-50 by shallow classifiers in order to classify images represented by bottleneck features. We evaluate the performance of proposed method replacing the top layer by three different classifiers: Support Vector Machine (SVM)~\cite{Bishop:2006:PRM:1162264}, k-Nearest Neighbor (kNN)~\cite{Bishop:2006:PRM:1162264}, and Extreme Gradient Boosting (XGBoost)~\cite{Chen_and_Guestrin_2016}.

For the SVM classifier we use two different kernels: a linear kernel, where the parameter $C$ has been obtained through a grid search process with $C \in \left[10^{-2},10^{-1},...,10^{10}\right]$, and a Radial Basis Function (RBF) kernel, where the parameters $C$ and $gamma$ ($\gamma$) have been obtained through a gridsearch process with $C \in \left[10^{-2},10^{-1},...,10^{10}\right]$ and $\gamma \in \left[10^{-9},10^{-8},...,10^{3}\right]$. 

The best $C$ obtained for linear kernel was $0.01$ and for RBF kernel the best $C$ obtained was $10.0$ with a $\gamma$ of $0.001$. Figure~\ref{fig:SVMgridsearch} shows the accuracies obtained in the gridsearch of parameters $C$ and $gamma$ using RBF kernel. For kNN classifier, we use a $k=1$ and for XGBoost the learning rate ($lr$) was $0.1$, maxdepth ($md$) was $3$ and the number of estimators ($ne$) was $100$.

\begin{figure}[h]
  \centering
  \includegraphics[width=.99\textwidth]{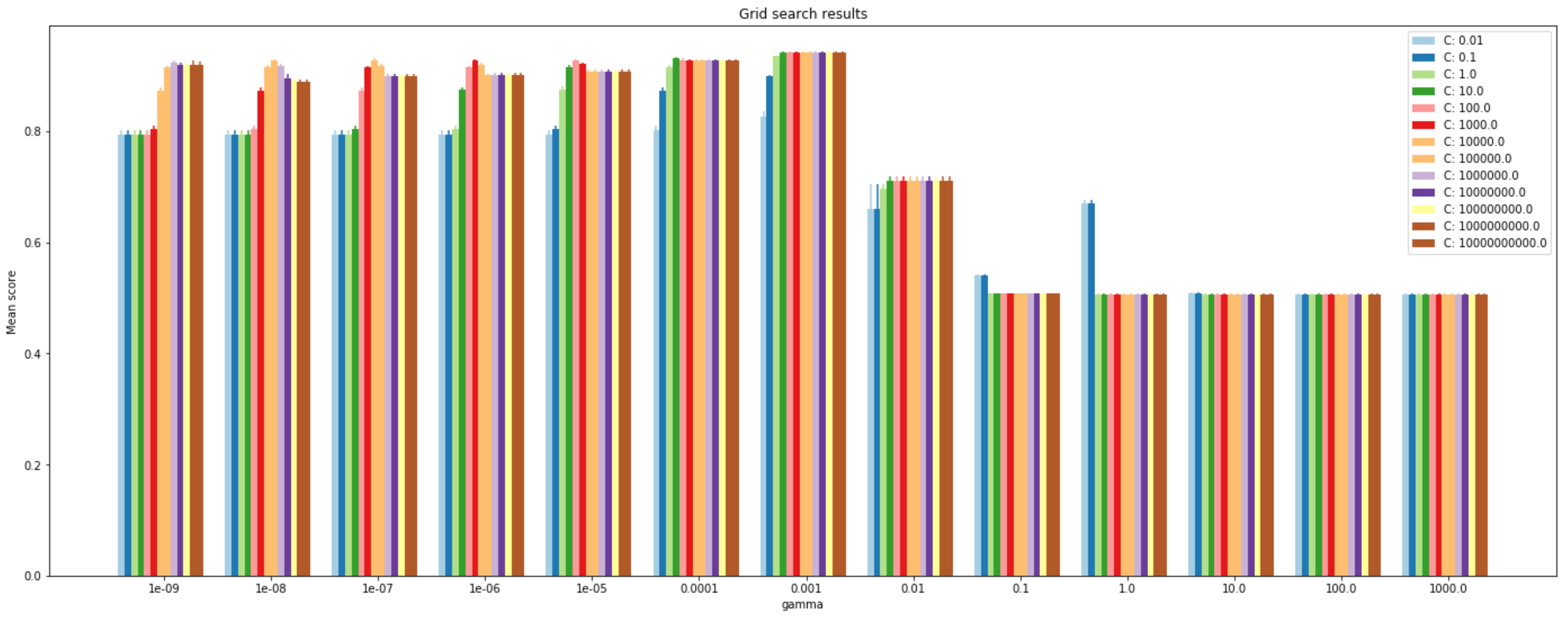}
  \caption{Gridsearch of parameters $C$ and $gamma$ of SVM with RBF kernel.}
  \label{fig:SVMgridsearch}
\end{figure}

Table~\ref{table:ResNet50-Classifiers-comparison} summarizes the results for each round of experiments, including the shallow classifiers proposed at this round. The best average accuracy achieved was $0.94$ using SVM with RBF kernel, with a standard deviation of $0.0065$ and a variance of $0.00003$.

\begin{table}[h]
\centering
\scriptsize
\caption{Summary of proposed approaches along 4 rounds of experiments.}
\label{table:ResNet50-Classifiers-comparison}
\begin{tabular}{|p{2.cm}|p{2cm}|p{1cm}|p{1.2cm}|p{0.7cm}|p{0.7cm}|p{1.5cm}|}
\hline
\rowcolor[HTML]{9B9B9B}
\textbf{Architecture} & \textbf{Train} & \textbf{Epochs} & \textbf{Transfer} & \textbf{Avg Acc} & \textbf{Std Dev} & \textbf{Variance}  \\ \hline
\rowcolor[HTML]{FFFFFF}
ResNet-50 + 2fc softmax & from scratch \qquad \qquad & 200 & no & 0.76 & 0.035 & 9.81E-04  \\ 
\rowcolor[HTML]{EFEFEF}
ResNet-50 + 2fc softmax & fine tune & 200 & yes & 0.82 & 0.035 & 9.73E-04  \\ 
\rowcolor[HTML]{FFFFFF}
ResNet-50 + 2fc softmax & pre-train top + fine tune & 200 & yes & 0.92 & 0.011 & 9.79E-05  \\ 
\rowcolor[HTML]{EFEFEF}
ResNet-50 + kNN & $k$=1&  & yes & 0.89 & 0.006 & 4.41E-05  \\ 
\rowcolor[HTML]{FFFFFF}
ResNet-50 + XGBoost & $lr$=0.1, $md$=3, $ne$=100 &  & yes & 0.90 & 0.007 & 3.56E-05  \\ 
\rowcolor[HTML]{EFEFEF}
ResNet-50 + SVM Linear & $C$=0.01 &  & yes & 0.92 & 0.007 & 4.39E-04  \\ 
\rowcolor[HTML]{FFFFFF}
{\color[HTML]{3531FF} \textbf{ResNet-50 + SVM RBF}} & {\color[HTML]{3531FF}\textbf{$C$=10, $\gamma$=0.001}} &  & {\color[HTML]{3531FF}\textbf{yes}} & {\color[HTML]{3531FF}\textbf{0.94}} & {\color[HTML]{3531FF}\textbf{0.007}} & {\color[HTML]{3531FF}\textbf{3.38E-05}}  \\ 
\hline
\end{tabular}
\end{table}

The ROC curve is presented in Figure~\ref{fig:rocdstok}. Additionally, in Figure~\ref{fig:SVMlearningcurve} we also provide the learning curve for the SVM with RBF Kernel

\begin{figure}[h]
  \centering
  \includegraphics[width=0.5\columnwidth]{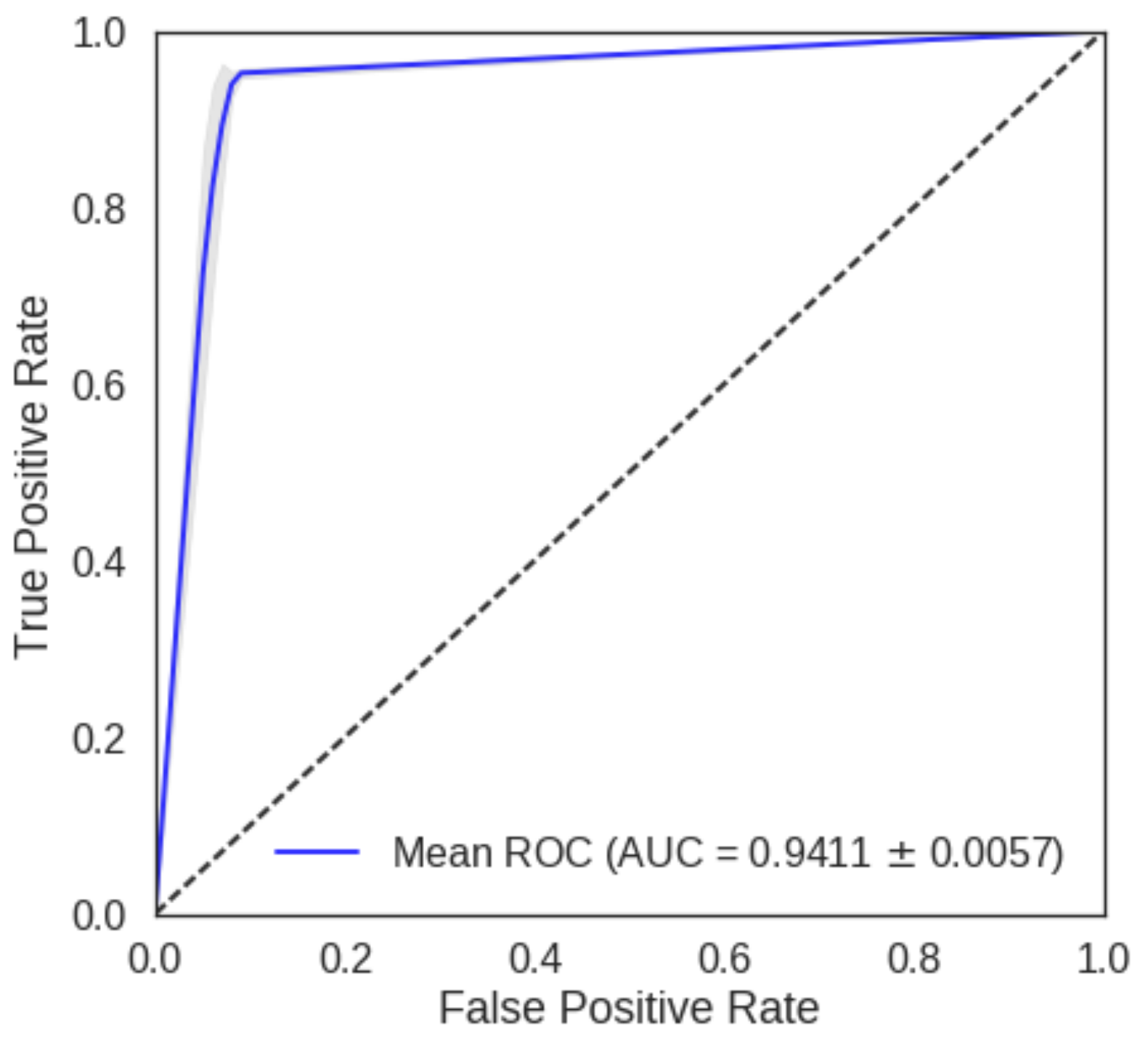}
  \caption{ROC curve for the best result in DSTok dataset using ResNet-50 + SVM RBF Kernel.}
  \label{fig:rocdstok}
\end{figure}

\begin{figure}[h]
  \centering
  \includegraphics[width=.90\textwidth]{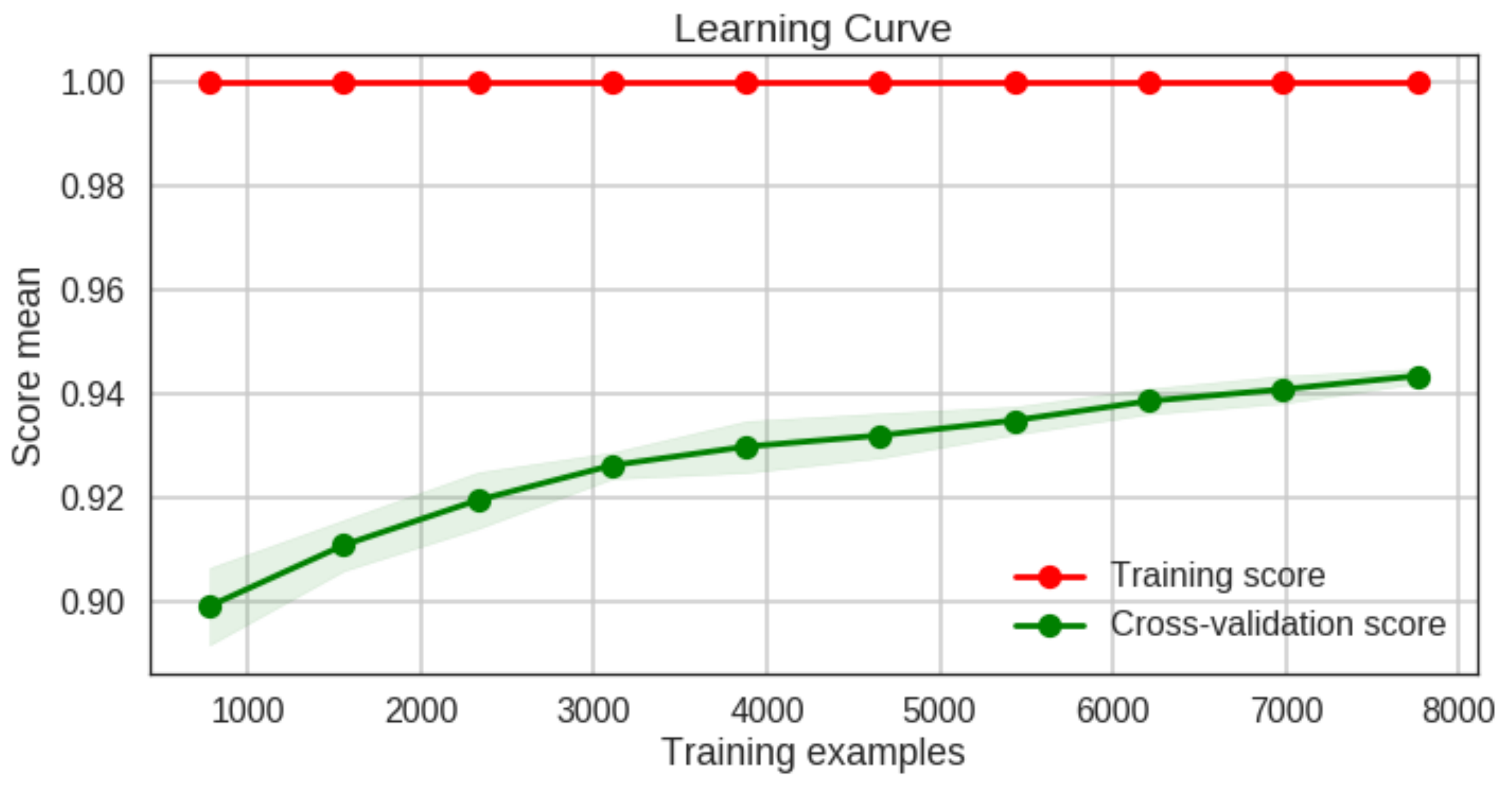}
  \caption{SVM Learning Curve.}
  \label{fig:SVMlearningcurve}
\end{figure}

Analyzing the learning curve, it is possible to observe that the training score is around the maximum and the validation score could be increased with more training samples. This observation led us to the next round of experiments, where we evaluate the transfer learning combined with SVM RBF performance in DSTokExt Dataset, which is an extended version of DSTok dataset proposed by \citeauthor{Tokuda20131276}~\cite{Tokuda20131276}.

\subsection{Round \#5: ResNet-50 bottleneck features with SVM over DSTokExt}
\label{sec:svmdstokext}

The transfer of ResNet-50 convolutional layers trained on ImageNet dataset to CG image detection problem provided results comparable to the best literature methods. Furthermore, as showed in Round \#4, the learning curve of ResNet 50 + SVM RBF suggests that increasing the number of samples could improve the model accuracy.

At this round of experiments, we use an extended version of DSTok dataset, named DSTokExt and described in Section~\ref{sec:dstokext}, to improve methods accuracy. We used the best model (ResNet-50 + SVM RBF) obtained in Round \#4 and performed a gridsearch to find the parameters $C$ and $gamma$ of the SVM RBF classifier with $C \in \left[10^{-2},10^{-1},...,10^{10}\right]$ and $\gamma \in \left[10^{-9},10^{-8},...,10^{3}\right]$. 
The best $C$ was $10.0$ with a $\gamma$ of $0.001$, the same values obtained in Round \#4. The average accuracy achieved was 0.97 with an standard deviation of 0.003 and a variance of 6.85E-06. The ROC curve is depicted in Figure~\ref{fig:SVMROCcurveext}. The area under the curve (AUC) is 0.97 and Table~\ref{table:ResNet50-SVM-DSTokExt} presents the accuracy for each fold.

\begin{figure}[h]
  \centering
  \includegraphics[width=0.5\columnwidth]{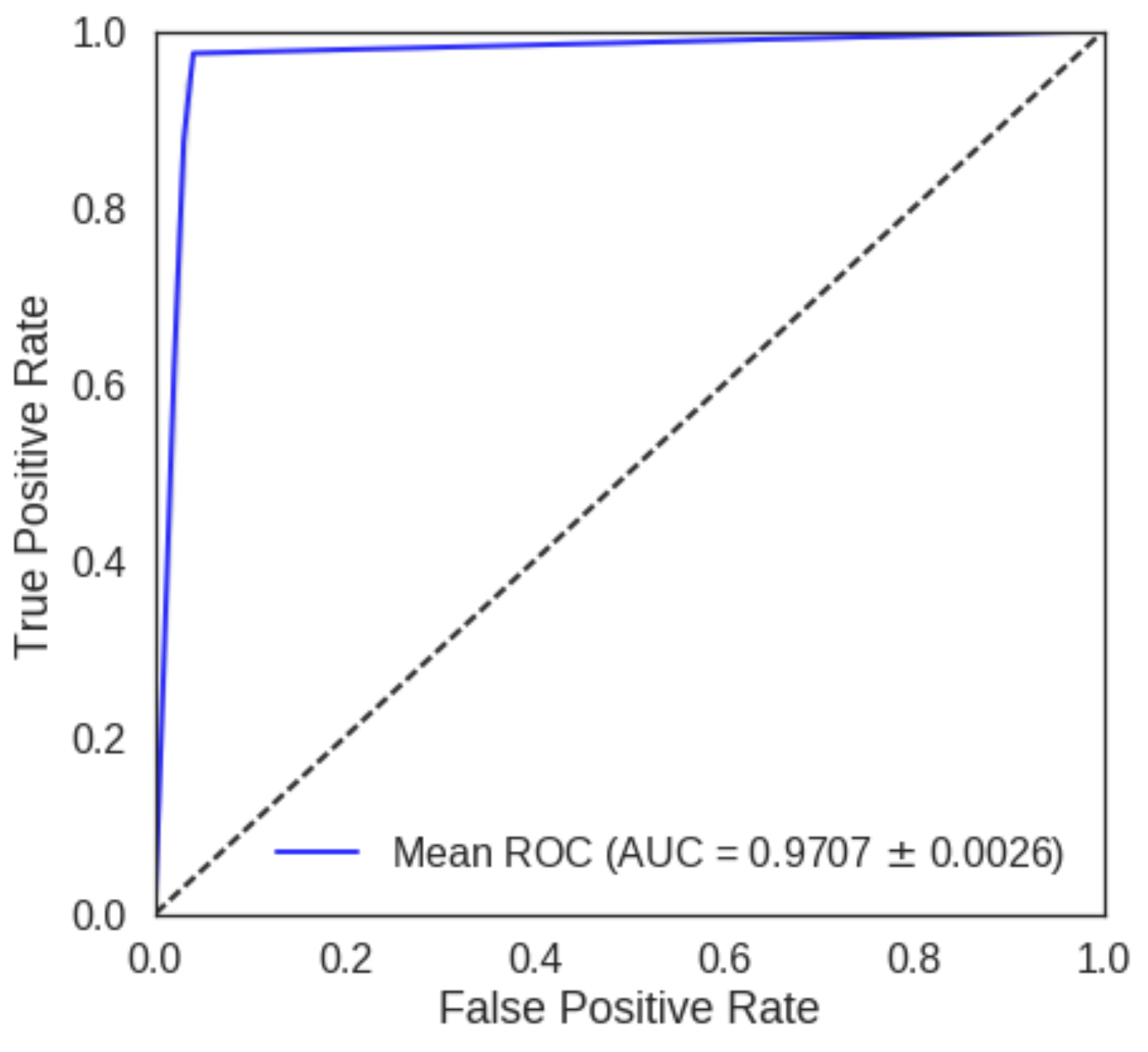}
  \caption{ROC curve for the best result in DSTokExt dataset using ResNet-50 + SVM RBF Kernel.}
  \label{fig:SVMROCcurveext}
\end{figure}

\begin{table}[h]
\centering
\scriptsize
\caption{Accuracy by fold of ResNet-50 + SVM RBF over DSTokExt.}
\label{table:ResNet50-SVM-DSTokExt}
\begin{tabular}{|p{1.2cm}|p{2cm}|}
\hline
\rowcolor[HTML]{9B9B9B}
\textbf{Fold}    & \textbf{Accuracy} \\ \hline
\rowcolor[HTML]{FFFFFF}
0       & 0.97   \\
\rowcolor[HTML]{EFEFEF}
1       & 0.98   \\
\rowcolor[HTML]{FFFFFF}
2       & 0.97   \\
\rowcolor[HTML]{EFEFEF}
3       & 0.97   \\
\rowcolor[HTML]{FFFFFF}
4       & 0.96   \\ 
\rowcolor[HTML]{EFEFEF}
{\color[HTML]{3531FF} \textbf{Average}}& {\color[HTML]{3531FF} \textbf{0.97}}  \\ \hline
\end{tabular}
\end{table}

Figure~\ref{fig:SVMlearningcurveDSTokExt} presents the learning curve for this round, with the red curve representing the training score, green curve representing the average test score and the green shadow representing the standard deviation across the folds. Again, as depicted in learning curve of Round \#4, it is possible to observe that the validation score could still be increased with more training samples.

\begin{figure}[h]
  \centering
  \includegraphics[width=.8\textwidth]{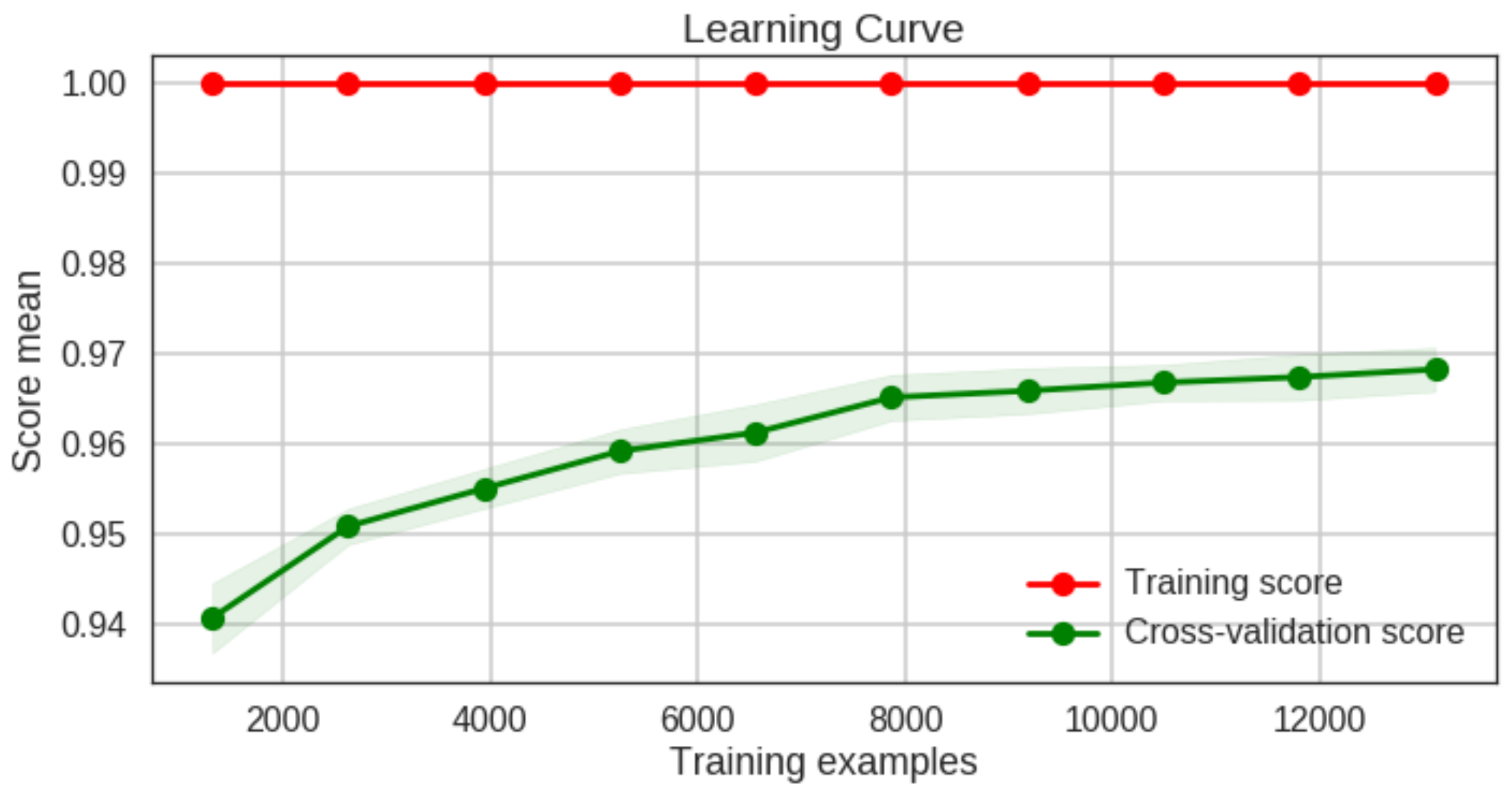}
  \caption{SVM Learning Curve over DSTokExt dataset.}
  \label{fig:SVMlearningcurveDSTokExt}
\end{figure}

Figure~\ref{fig:SVMConfusionMatrix} shows a comparison of confusion matrix (left) and the normalized confusion matrix (right) obtained with ResNet-50 + SVM RBF on DSTok and DSTokExt datasets.

\begin{figure}[h]
  \centering
  \subfloat[DSTok dataset]{\includegraphics[width=0.3\columnwidth]{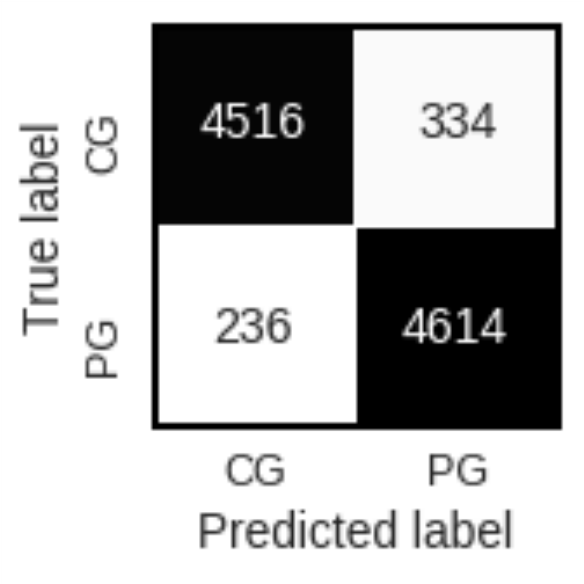} \hspace{0.01\columnwidth} \includegraphics[width=0.3\columnwidth]{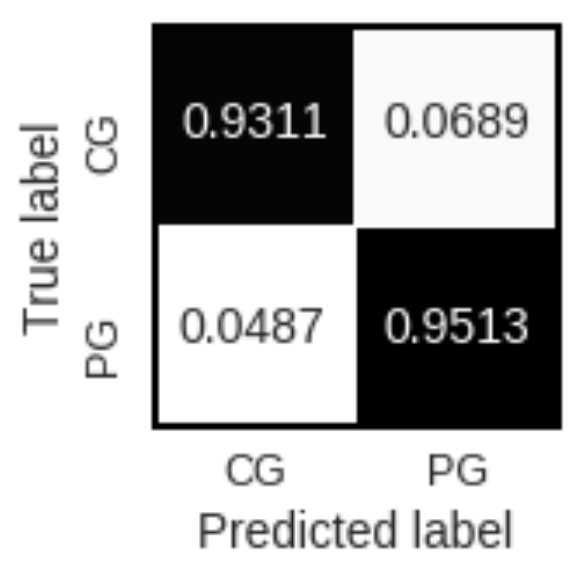}}
  \\
  \subfloat[DSTokExt dataset]{\includegraphics[width=0.3\columnwidth]{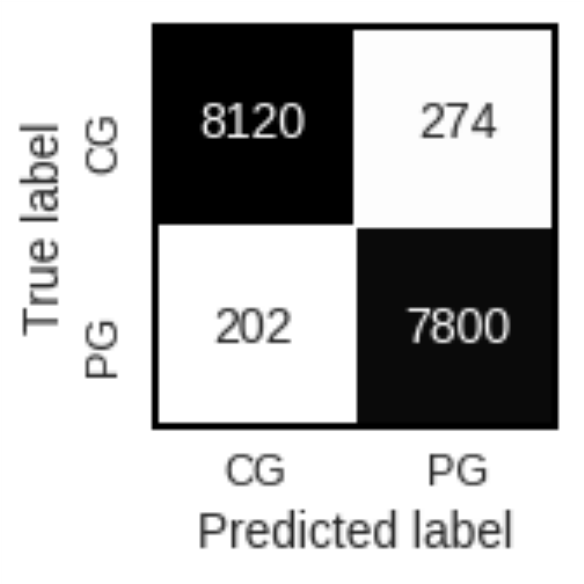} \hspace{0.01\columnwidth} \includegraphics[width=0.3\columnwidth]{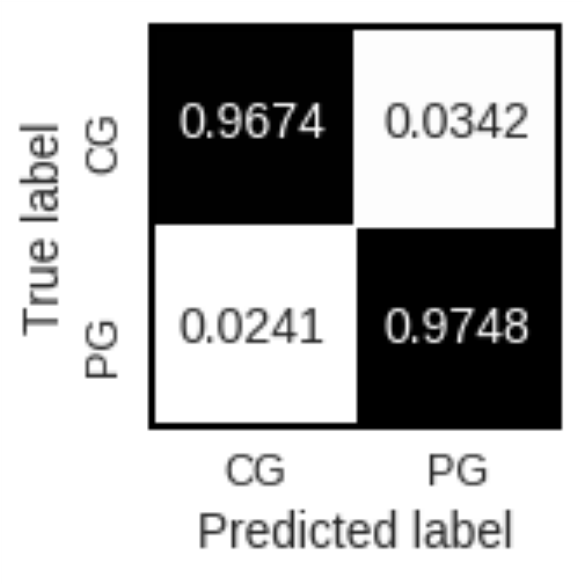}}
  \caption{Confusion matrix of the SVM classifier.}
  \label{fig:SVMConfusionMatrix}
\end{figure}

\subsection{Round \#6: Visualization of Bottleneck Features}
\label{sec:visualizationbottleneck}

As described in Section~\ref{sec:proposedmethod}, our method takes advantage of transfer learning process to generate ResNet-50 bottleneck features, projecting the 150,528 input features ($224\times224\times3$ RGB values of the pixels of each image) in a lower-dimensional space of 2,048 features. This process intends to generate a set of features with a better degree of separability, which could allow the top classifier to achieve a higher classification accuracy.

To evaluate if the bottleneck features would, in fact, produce the desired boost in classification accuracy, we applied the t-Distributed Stochastic Neighbor Embedding (t-SNE)~\cite{maaten2008visualizing} dimensionality reduction technique to visualize our high-dimensional features. We projected the 150,528 input features and the 2,048 bottleneck features in 2D, and plot them as points colored according to their class, as depicted in Figure~\ref{fig:tSNE-DSTok} for the images in DSTok dataset and Figure~\ref{fig:tSNE-DSTokExt} for the images in DSTokExt dataset. Green circles represent CG samples while blue squares represent PG samples.

\begin{figure}[h]
  \centering
  \subfloat[Raw image pixels]{\includegraphics[width=0.49\columnwidth]{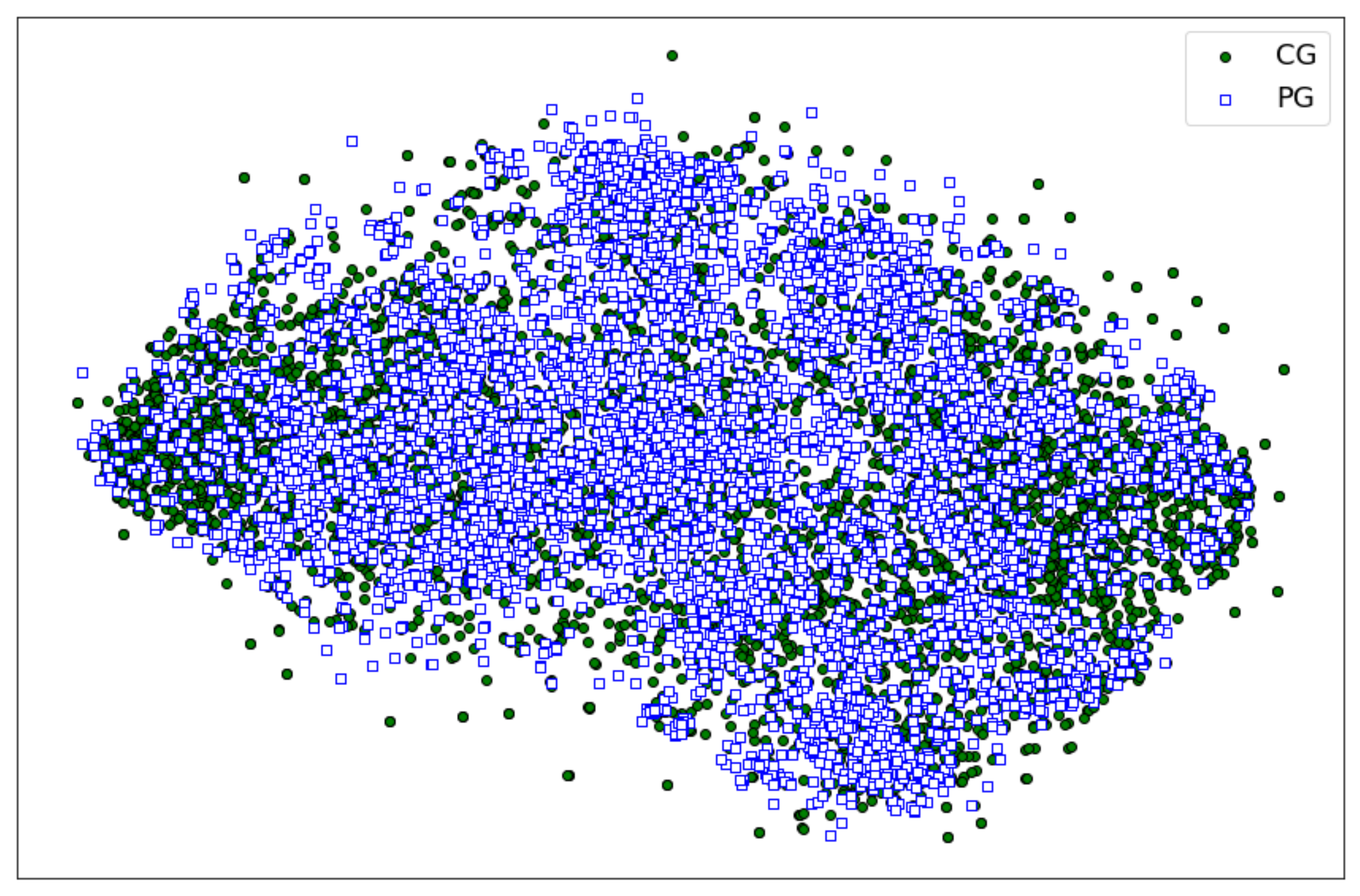}}
  \subfloat[Bottleneck features]{\includegraphics[width=0.49\columnwidth]{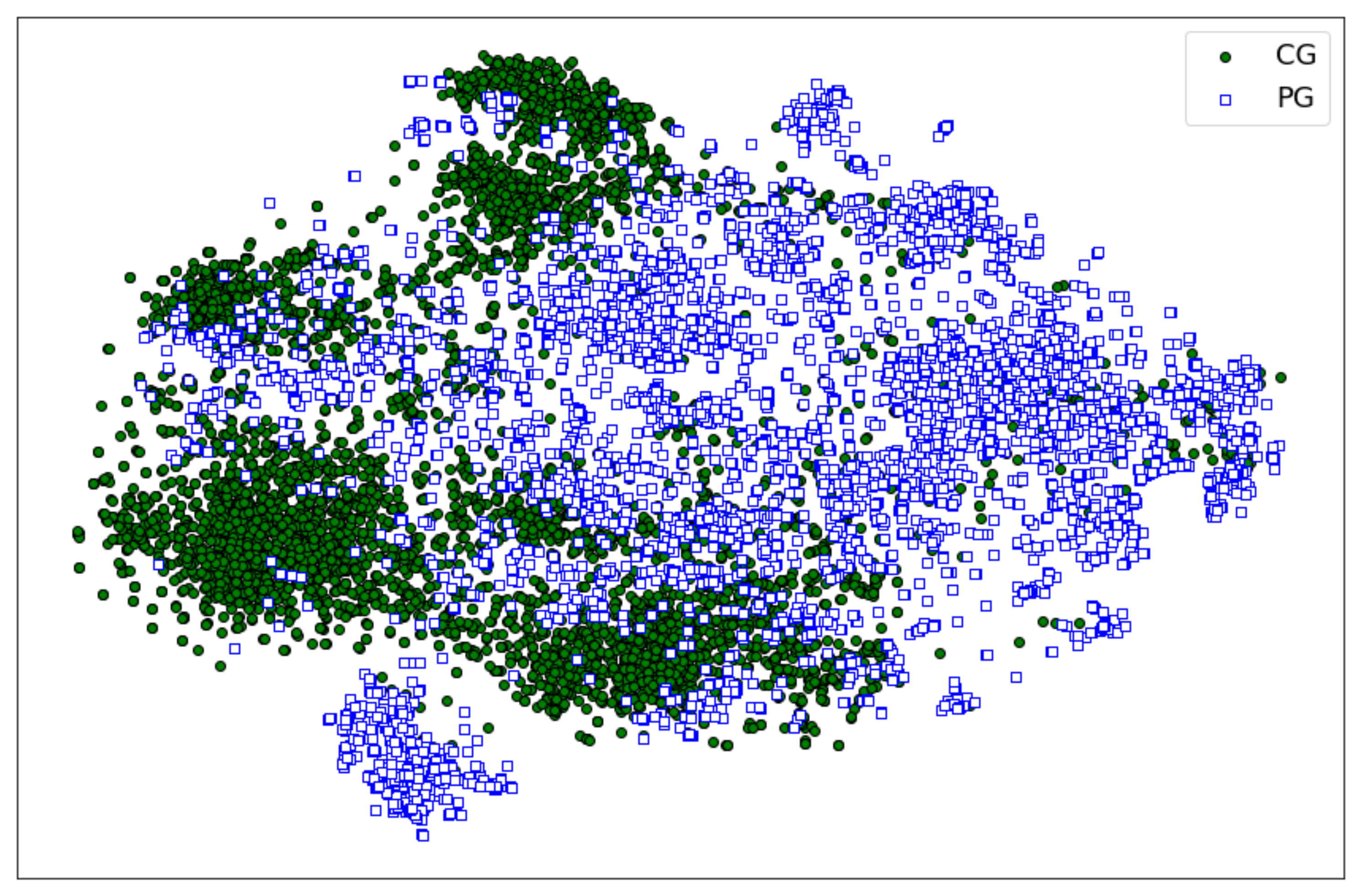}}
  \caption{t-SNE visualization of DSTok dataset using (a) raw image pixels and (b) ResNet-50 bottleneck features.}
 \label{fig:tSNE-DSTok}
\end{figure}

\begin{figure}[h]
  \centering
  \subfloat[Raw image pixels]{\includegraphics[width=0.49\columnwidth]{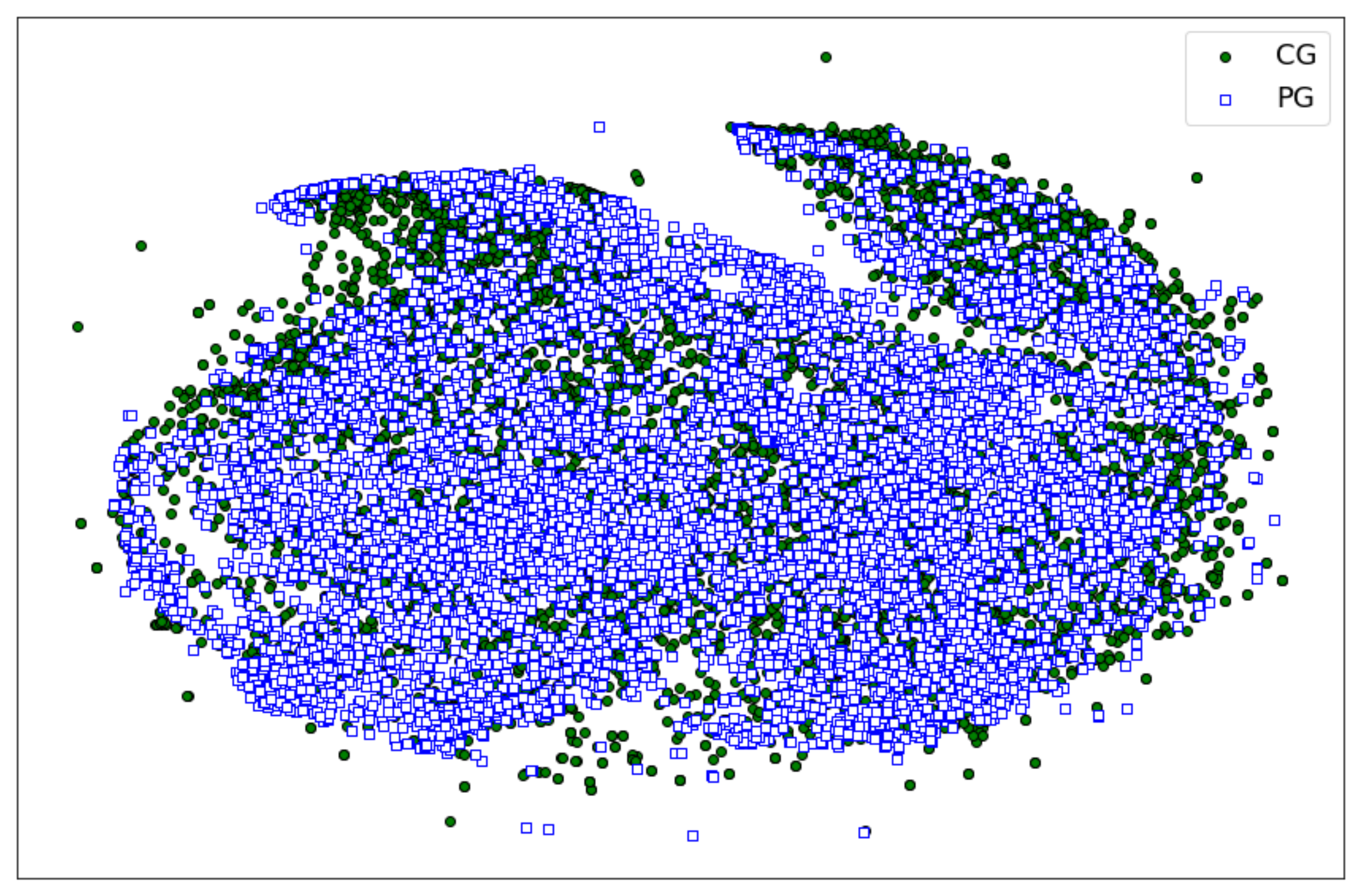}}
  \subfloat[Bottleneck features]{\includegraphics[width=0.49\columnwidth]{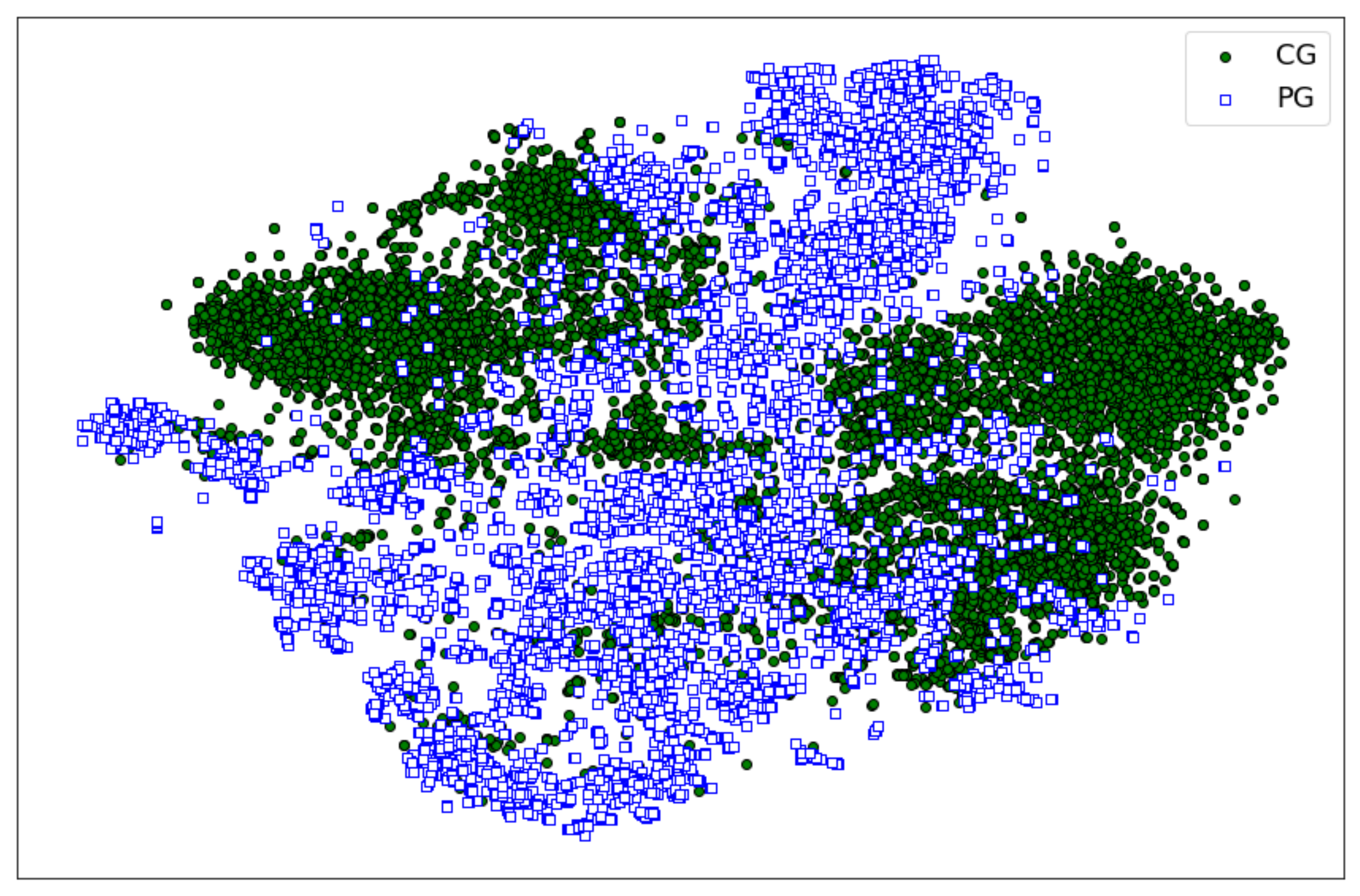}}
  \caption{t-SNE visualization of DSTokExt dataset using (a) raw image pixels and (b) ResNet-50 bottleneck features.}
 \label{fig:tSNE-DSTokExt}
\end{figure}

It is possible to observe in the figures that the operations performed by ResNet-50 convolutional layers projected the raw pixels into a better separable feature space.

\subsection{Round \#7:Comparative Analysis}

Along five rounds of experiments, we exposed how transfer learning can be used to take advantage of a DNN trained for object recognition task to generate discriminative features for CG images. These features can be used to train accurate classifiers for CG detection problem. The best accuracy of 0.97 was achieved using bottleneck features with an SVM classifier with RBF kernel in an extended version of DSTok dataset (containing DSTok images plus additional images).

In \etal{Tokuda}~\cite{Tokuda20131276}, the authors present an extensive comparison of several literature approaches dedicated to solve the problem of detecting CG and PG images. The main characteristics of each method investigated by the authors are reported in Table~\ref{table:Methods}. Additionally, we included the characteristics of all methods proposed in this work: (1) ResNet-50 trained from Glorot uniform initialization (DNN1); (2) ResNet-50 fine-tuned from ImageNet initialization (DNN2); (3) ResNet-50 fine-tuned from ImageNet initialization and pre-trained softmax layer (DNN3); (4) ResNet-50 bottleneck features with kNN (DNN4); (5) ResNet-50 bottleneck features with XGBoost (DNN5); (6) ResNet-50 bottleneck features with SVM Linear (DNN6); and (7) ResNet-50 bottleneck features with SVM RBF (DNN7).

\begin{table}[h]
\centering
\scriptsize
\caption{Methods evaluated by \etal{Tokuda}~\cite{Tokuda20131276} and methods proposed here. For each of one of the methods, it is shown the identifiers, the main concepts and the related features used by the methods.}
\label{table:Methods}
\begin{tabular}{|p{1.7cm}|p{8cm}|p{2.5cm}|}
\hline
\rowcolor[HTML]{9B9B9B}
\textbf{Method} & \textbf{Main Concept} & \textbf{Feature}\\ \hline
\rowcolor[HTML]{FFFFFF}
Li~\cite{5569821}     & Second order differences      & Edges/Texture\\ 
\rowcolor[HTML]{EFEFEF}
LSB~\cite{ng2009}    & Camera noise                  & Acquisition   \\
\rowcolor[HTML]{FFFFFF}
LYU~\cite{Lyu2005}    & Wavelet transform             & Edges/Texture \\
\rowcolor[HTML]{EFEFEF}
POP~\cite{Popescu2005}    & Interpolator predictor        & Acquisition   \\
\rowcolor[HTML]{FFFFFF}
BOX~\cite{Liebovitch1989}    & Boxes counting               & Auto-similarity \\
\rowcolor[HTML]{EFEFEF}
CON~\cite{vetterli2002}     & Contourlet transform         & Edges/Texture \\
\rowcolor[HTML]{FFFFFF}
CUR~\cite{candes2000}    & Curvelet transform~\cite{candes2000}            & Edges/Texture \\
\rowcolor[HTML]{EFEFEF}
GLC~\cite{Haralick1973}    & Cooccurrence matrix           & Texture       \\
\rowcolor[HTML]{FFFFFF}
HOG~\cite{triggs2005}    & Histogram of oriented grads  & Shape         \\
\rowcolor[HTML]{EFEFEF}
HSC~\cite{schwartz1056}    & Histogram of shearlet coeff  & Curves        \\
\rowcolor[HTML]{FFFFFF}
LBP~\cite{ojala2001}    & Local binary patterns         & Edges/Texture \\
\rowcolor[HTML]{EFEFEF}
SHE~\cite{Kutyniok2011}     & Shearlet transform           & Edges/Texture\\
\rowcolor[HTML]{FFFFFF}
SOB~\cite{gonzales2007}    & Sobel operator                & Edges         \\
\rowcolor[HTML]{EFEFEF}
FUS1~\cite{Tokuda20131276}   & Concatenation                 & Combination   \\
\rowcolor[HTML]{FFFFFF}
FUS2~\cite{Tokuda20131276}   & Simple voting                 & Combination   \\
\rowcolor[HTML]{EFEFEF}
FUS3~\cite{Tokuda20131276}   & Weighted voting               & Combination   \\
\rowcolor[HTML]{FFFFFF}
FUS4~\cite{Tokuda20131276}   & Meta-classification           & Combination   \\
\rowcolor[HTML]{EFEFEF}
DNN1    & Deep CNN + Softmax (from scratch)   & Raw image pixels\\
\rowcolor[HTML]{FFFFFF}
DNN2    & Deep CNN transfer + Softmax (from ImageNet weights)       & Raw image pixels\\ 
\rowcolor[HTML]{EFEFEF}
DNN3    & Deep CNN transfer + Softmax (fine-tuning)   & Raw image pixels\\
\rowcolor[HTML]{FFFFFF}
DNN4    & Deep CNN transfer + kNN & Raw image pixels\\
\rowcolor[HTML]{EFEFEF}
DNN5    & Deep CNN transfer + XGBoost & Raw image pixels\\
\rowcolor[HTML]{FFFFFF}
DNN6    & Deep CNN transfer + SVM Linear & Raw image pixels\\
\rowcolor[HTML]{EFEFEF}
DNN7    & Deep CNN transfer + SVM RBF & Raw image pixels\\
\hline
\end{tabular}
\end{table}

Considering that our experimental protocol is exactly the same one adopted by \etal{Tokuda}~\cite{Tokuda20131276}, we used the results reported by the authors to compare our method with other literature methods. In addition, we also include the results obtained using the DSTokExt dataset in the comparison table. Table~\ref{table:Comparison} presents these results. From the table, we see that the accuracies of literature methods have a large range of values going from 0.97 (highest) to 0.552 (lowest). Proposed method DNN7 overcome all literature methods based on raw and simple features and it is better than FUS1 proposed by \citeauthor{Tokuda20131276}~\cite{Tokuda20131276}. This fact shows the expression power of transfer learning approach in features extraction process. Additionally, when DNN7 have the number of training samples increased, it achieves the same accuracy as the best approach proposed by \citeauthor{Tokuda20131276}~\cite{Tokuda20131276} but with a lower variance.

\begin{table}[h]
\centering
\scriptsize
\caption{Comparison among approaches for distinguishing CGs and PGs. Table is sorted from highest to lowest average accuracy. For each of the methods, it is shown the number of dimensions of the feature space (m), the average accuracy for each class, the variance and the dataset where the experiments have been performed.}
\label{table:Comparison}
\begin{tabular}{|p{1.4cm}|p{1cm}|p{1.4cm}|p{1.4cm}|p{2.5cm}|}
\hline
\rowcolor[HTML]{9B9B9B}
\textbf{Method} & \textbf{m}    &  \textbf{Average accuracy} & \textbf{Variance} & \textbf{Dataset} \\
\hline
\rowcolor[HTML]{FFFFFF}
\textbf{DNN7}    & \textbf{150528} & \textbf{0.97} & \textbf{6.85E-06} & \textbf{DSTokExt} \\
\rowcolor[HTML]{EFEFEF}
FUS4   & 13     & 0.97 & 6.06E-04 & DSTok \\
\rowcolor[HTML]{FFFFFF}
FUS3   & 13   & 0.96 & 3.86E-04 & DSTok \\
\rowcolor[HTML]{EFEFEF}
FUS2   & 13   & 0.95 & 2.82E-04 & DSTok \\
\rowcolor[HTML]{FFFFFF}
\textbf{DNN7}    & \textbf{150528} & \textbf{0.94} & \textbf{3.38E-05} & \textbf{DSTok} \\
\rowcolor[HTML]{EFEFEF}
FUS1   & 4011 & 0.93 & 9.60E-02 & DSTok \\
\rowcolor[HTML]{FFFFFF}
Li     & 144    & 0.93 & 8.27E-05 & DSTok \\
\rowcolor[HTML]{EFEFEF}
\textbf{DNN6}    & \textbf{150528} & \textbf{0.92} & \textbf{4.39E-04} & \textbf{DSTok} \\
\rowcolor[HTML]{FFFFFF}
\textbf{DNN3}    & \textbf{150528} & \textbf{0.92} & \textbf{9.79E-05} & \textbf{DSTok} \\
\rowcolor[HTML]{EFEFEF}
LYU    & 216    & 0.92 & 2.26E-04 & DSTok \\
\rowcolor[HTML]{FFFFFF}
\textbf{DNN5}    & \textbf{150528} & \textbf{0.90} & \textbf{3.56E-05} & \textbf{DSTok} \\
\rowcolor[HTML]{EFEFEF}
CON    & 696    & 0.90 & 3.03E-04 & DSTok \\
\rowcolor[HTML]{FFFFFF}
\textbf{DNN4}    & \textbf{150528} & \textbf{0.89} & \textbf{4.41E-05} & \textbf{DSTok} \\
\rowcolor[HTML]{EFEFEF}
LBP    & 78     & 0.87 & 3.68E-04 & DSTok \\
\rowcolor[HTML]{FFFFFF}
\textbf{DNN2}    & \textbf{150528} & \textbf{0.82} & \textbf{9.73E-04} & \textbf{DSTok} \\
\rowcolor[HTML]{EFEFEF}
CUR    & 2328   & 0.80& 9.39E-04 & DSTok \\
\rowcolor[HTML]{FFFFFF}
HSC    & 96     & 0.80 & 6.23E-04 & DSTok \\
\rowcolor[HTML]{EFEFEF}
\textbf{DNN1}    & \textbf{150528} & \textbf{0.76} & \textbf{9.81E-04} & \textbf{DSTok} \\
\rowcolor[HTML]{FFFFFF}
HOG    & 256    & 0.74 & 5.20E-04 & DSTok \\
\rowcolor[HTML]{EFEFEF}
SHE    & 60     & 0.71 & 7.84E-04 & DSTok \\
\rowcolor[HTML]{FFFFFF}
LSB    & 12     & 0.66 & 7.53E-04 & DSTok \\
\rowcolor[HTML]{EFEFEF}
GLC    & 12     & 0.63 & 1,01E-03 & DSTok \\
\rowcolor[HTML]{FFFFFF}
POP    & 12     & 0.57 & 5.95E-04 & DSTok \\
\rowcolor[HTML]{EFEFEF}
BOX    & 3      & 0.55 & 1.45E-03 & DSTok \\
\rowcolor[HTML]{FFFFFF}
SOB    & 150    & 0.55 & 1,05E-03 & DSTok \\
\hline
\end{tabular}
\end{table}

In Table~\ref{table:Comparison}, it is possible to observe that our method performs better than the lowest fusion approach, even using a single kind of feature. Moreover, increasing the size of training dataset, our approach presents the same result as the best fusion proposed by \citeauthor{Tokuda20131276}~\cite{Tokuda20131276} with two main advantages: the absence of laborious hand-craft feature extraction work and a lower variance in results (showing more stability).
\section{Conclusions and Research Directions}
\label{sec:conclusions}

In this paper we have presented a new method for CG images detection using a deep convolutional neural network model based on ResNet-50 and transfer learning concepts. After a simple pre-processing, each image in our dataset is fed into our deep CNN model and, as result, we obtain a 2048 dimension feature vector, here called bottleneck features. Exploring different approaches looking for achieving the most effective problem solution, we evaluate different approaches, since train ResNet-50 architecture from scratch (just changing the 1000fc softmax from original architecture to a 2fc softmax in top layer), using our dataset, for CG detection process, until full transfer learning, where ImageNet weights for ResNet-50 are totally frozen in a way to produce bottleneck features, which are used to train different machine learning classifiers to detect if an image is, or not, produced by computer graphics methods.

Conducting different rounds of experiments, we evaluate the efficiency and effectiveness of using a Deep CNN architecture, proposed for a object recognition task, in a CG detection problem, where we are looking for distinguish between a CG and a PG image, involving different kinds of objects and context. Results showed that proposed approach perform as good as state-of-the-art methods in the same dataset, achieving more than 0.97 accuracy rate. These results highlight two main advantages of proposed method: (1) absence of require a hand-craft feature extraction and (2) more stability depicted by a lower variance.

In special, in Round \#6 of experiments (Section~\ref{sec:visualizationbottleneck}) we showed for DSTok dataset, using t-SNE dimensionality reduction, the expression power of bottleneck features generated by ResNet-50 transfer layers, which increases the classes separability when compared against raw input features. The same behavior is kept in DSTokExt with more and different kinds of images.

Furthermore, it is important to realize that, as showed in Section~\ref{sec:svmdstokext}, even with a extended dataset, the learning curve from the SVM classifier it is still not stable, which suggest that training score is not still around the maximum. Since deep learning models needs an astonishing number of images to achieve a satisfactory accuracy, we conclude that keeps increasing the number of images can leads to an accuracy even better.

A limitation of this method is its difficulty in dealing with CG images with a high degree of realism. We conducted an experiment with two datasets involving images very similar and with high degree of realism, one proposed by \citeauthor{holmes_etal}~\cite{holmes_etal} and a second one proposed by \citeauthor{carvalhoicml2017}~\cite{carvalhoicml2017}. This left a door open for the improvement of this technique or development of a new one that could cope with this difficult scenario.

As research directions, our propose is explore different architectures bottleneck features extraction and perform fusion of these architectures in a way to construct an ensemble of deep architectures.

\section*{Acknowledgments}

The authors would like to thank the financial support of IFSP-Campinas, FAPESP (grant 2017/12631-6) and CNPq (grants 302923/2014-4, 313152/2015-2 and 423797/2016-6). We also would like to thank the authors \etal{Tokuda}~\cite{Tokuda20131276} who helped us with dataset acquirement and we gratefully acknowledge the support of NVIDIA Corporation with the donation of the GPUs used for this research.

\clearpage

\section*{References}
\bibliographystyle{elsarticle-num-names}
\bibliography{references}

\end{document}